\documentclass[lettersize,journal]{IEEEtran}
\usepackage{amsmath,amsfonts}
\usepackage{array}
\usepackage[caption=false,font=normalsize,labelfont=sf,textfont=sf]{subfig}
\usepackage{textcomp}
\usepackage{stfloats}
\usepackage{url}
\usepackage{verbatim}
\usepackage{graphicx}
\usepackage[T1]{fontenc}
\usepackage{threeparttable}
\usepackage[numbers,sort&compress]{natbib}
\def\BibTeX{{\rm B\kern-.05em{\sc i\kern-.025em b}\kern-.08em
    T\kern-.1667em\lower.7ex\hbox{E}\kern-.125emX}}
\usepackage{balance}
\usepackage{multirow}
\usepackage{utfsym}
\usepackage{hyperref}
\usepackage{float}
\usepackage{color}
\usepackage{xcolor}
\usepackage{colortbl}
\usepackage{algorithm}
\usepackage{algpseudocode}
\usepackage{bm}
\usepackage{amsmath}
\usepackage{algorithmicx}
\usepackage{algpseudocode}
\usepackage{tabularx}
\hypersetup{
	colorlinks,
	linkcolor=blue, 
	citecolor=cyan!50!gray,
	urlcolor=cyan
}

\begin{document}
\title{AsyReC: A Multimodal Graph-based Framework for Spatio-Temporal Asymmetric Dyadic Relationship Classification}

\author{Wang Tang, Fethiye Irmak Dogan, Linbo Qing, Hatice Gunes, \textit{Senior Member}, \textit{IEEE}
\thanks{Wang Tang, Fethiye Irmak Dogan and Hatice Gunes are with the AFAR Lab, Department of Computer Science and Technology, University of Cambridge, CB3 0FT Cambridge, U.K. (e-mail:\{wt299,fid21,hg410\}@cam.ac.uk). This research was undertaken while Wang Tang was a visiting PhD student at the Cambridge AFAR Lab.
\noindent\textbf{Funding:} T. Wang is supported by China Scholarship Council (CSC). F. I. Dogan and H. Gunes are supported by the EPSRC/UKRI under grant ref. EP/R030782/1 (ARoEQ).
\textbf{Open Access:} For open access purposes, the authors have applied a Creative Commons Attribution (CC BY) licence to any Author Accepted Manuscript version arising. \textbf{Data access:} This work undertakes secondary analyses on existing datasets that are cited accordingly in the paper.
}
\thanks{Wang Tang and Linbo Qing are with the College of Electronics and Information Engineering, Sichuan University, Chengdu 610065, China (e-mail:tangwang@stu.scu.edu.cn, qing$\_$lb@scu.edu.cn). \noindent\textbf{Funding:} Wang Tang and Linbo Qing are supported by the Xizang Key Research and Development Program under grant No. XZ202501ZY0064.
}
}


\maketitle
\begin{abstract}
Dyadic social relationships, which refer to relationships between two individuals who know each other through repeated interactions (or not), are shaped by shared spatial and temporal experiences. Current computational methods for modeling these relationships face three major challenges: (1) the failure to model asymmetric relationships, e.g., one individual may perceive the other as a \emph{friend} while the other perceives them as an \emph{acquaintance}, (2) the disruption of continuous interactions by discrete frame sampling, which segments the temporal continuity of interaction in real-world scenarios, and (3) the limitation to consider periodic behavioral cues, such as rhythmic vocalizations or recurrent gestures, which are crucial for inferring the evolution of dyadic relationships. To address these challenges, we propose AsyReC, a multimodal graph-based framework for asymmetric dyadic relationship classification, with three core innovations: (i) a triplet graph neural network with node-edge dual attention that dynamically weights multimodal cues to capture interaction asymmetries (addressing challenge 1); (ii) a clip-level relationship learning architecture that preserves temporal continuity, enabling fine-grained modeling of real-world interaction dynamics (addressing challenge 2); and (iii) a periodic temporal encoder that projects time indices onto sine/cosine waveforms to model recurrent behavioral patterns (addressing challenge 3). Extensive experiments on two public datasets demonstrate state-of-the-art performance, while ablation studies validate the critical role of asymmetric interaction modeling and periodic temporal encoding in improving the robustness of dyadic relationship classification in real-world scenarios. Our code is publicly available at: \href{https://github.com/tw-repository/AsyReC}{https://github.com/tw-repository/AsyReC}.
\end{abstract}


\section{Introduction}
\IEEEPARstart{H}{uman} social relationships emerge from the sum of the social interactions between individuals over a period of time~\cite{38}, characterized by complex verbal and nonverbal cues, such as conversational dynamics and body language, which provide critical insights into relationship contexts~\cite{02}. Computational modeling of these interaction dynamics enables various applications, such as affective computing systems to analyse contextualised expresser-observer social interactions for understanding emotion dynamics~\cite{95}; socially intelligent robots to interpret interaction cues for context-aware responses~\cite{04}; and urban perception studies to analyse how social relationships shape different social groups~\cite{91}. Integrating social intelligence into computational systems holds significant potential for developing adaptive AI and socially aware environments to enhance the human experience~\cite{90}.

\begin{figure}[t!]
    \setlength{\abovecaptionskip}{0.cm}
    \setlength{\belowcaptionskip}{-0.cm}
    \centering
    \includegraphics[width=\linewidth]{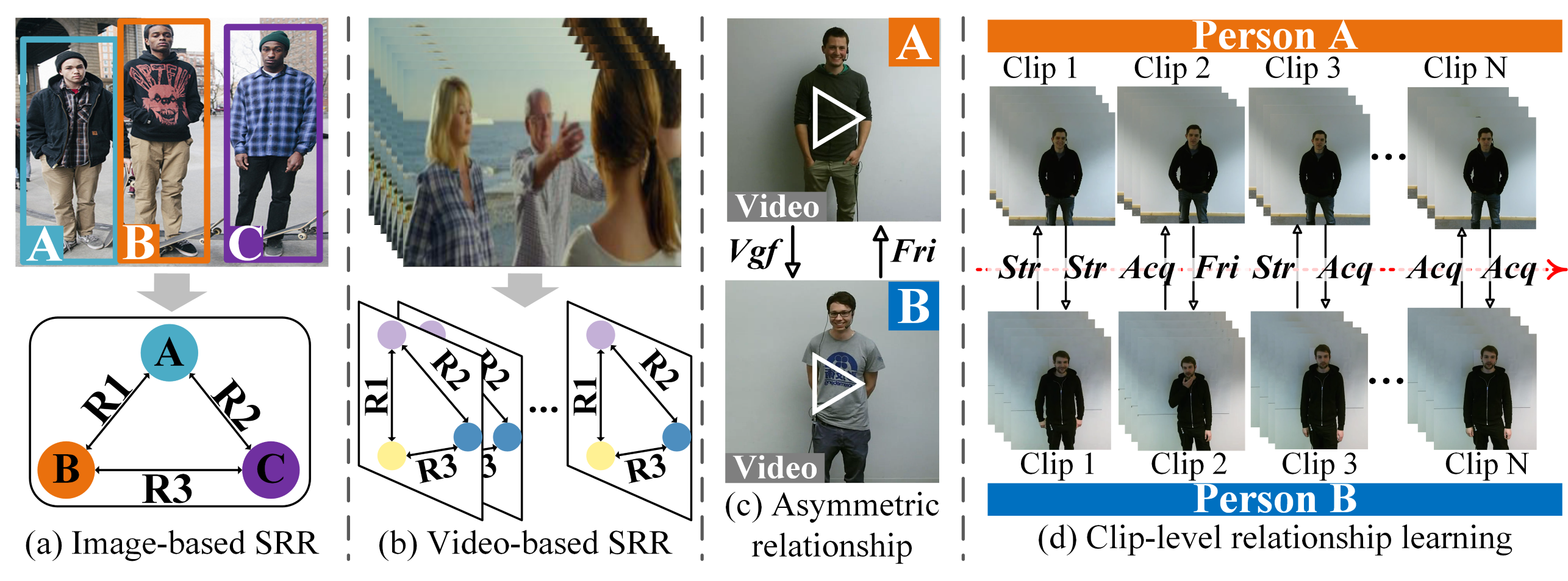}
    \vspace{-1.0em}
    \caption{Existing research paradigms: (a) Image-based SRR with an image from the PISC~\cite{11} dataset. (b) Video-based SRR with a video from the ViSR~\cite{21} dataset. (c) Asymmetric relationship, where the relationship from person A's perspective is different from that of person B. (d) Clip-level relationship learning. The screenshots in (c) and (d) are from the NoXi database collected for the ARIA-VALUSPA project~\cite{49}. \textit{The abbreviations are Very good friend (Vgf), Friend (Fri), Stranger (Str), Acquaintance (Acq).} \label{Fig: 1}}
    \vspace{-1.0em}
\end{figure}

Current research in \textbf{S}ocial \textbf{R}elationship \textbf{R}ecognition (SRR) can be divided into two distinct paradigms: image-based and video-based approaches. Early work primarily focused on analysing relationships within static images~\cite{10,11}, leveraging visual features and spatial coordinates to predict relationships between individuals. Subsequent advances, illustrated in Fig.~\ref{Fig: 1}(a), introduced graph-structured representations to encode individual features for relationship inference ~\cite{13,15,16,17,18,19,20}. However, such image-based approaches inherently lack temporal information, a critical limitation that fails to capture the temporal dynamics in real-world social interactions, thereby limiting their ability to capture evolving interpersonal cues.

To address these limitations, video-based approaches have been developed to model interpersonal interactions by integrating spatio-temporal information~\cite{21,23}. As illustrated in Fig.~\ref{Fig: 1}(b), these methods construct a spatial relationship graph structure for certain key frames and subsequently fuse them to form a spatio-temporal graph inference framework. However, these approaches often overlook the inherent asymmetries in interpersonal dynamics. For instance (Fig.~\ref{Fig: 1}(c)), individual A may interpret individual B's rigid facial expression and closed posture as indicating a formal ``\textit{\textbf{Fri}end}'' (Fri) relationship, whereas B perceives A's sustained eye contact, relaxed posture, and warm vocal tone as signaling a ``\textit{\textbf{V}ery \textbf{g}ood \textbf{f}riend}'' (Vgf) relationship. Such asymmetries arise from differences in subjective experience and social roles~\cite{33,34}, which current methodologies fail to capture. Modeling these asymmetric patterns is challenging yet important for capturing perceptual divergences and elucidating the underlying behavioural patterns of interacting individuals (\textit{Challenge~1}).

Recent studies have also investigated multimodal fusion techniques~\cite{24,25,26,27,28,29}, which integrate visual, auditory, and textual features to identify social relationships, such as ``\textit{couple}'' or ``\textit{colleague}''. These approaches focus on relationship recognition in short videos, typically sourced from movies or television series~\cite{21,24,25}. For example, Fig.~\ref{Fig: 1}(b) depicts a scene from a movie, while Figs.~\ref{Fig: 1}(c) and (d) illustrate real-world human interactions. Movies and TV series benefit from structured narratives, which provide contextual guidance for deep learning models to understand the overarching storyline and the dynamic evolution of character relationships. In contrast, real-world interpersonal interactions lack such predefined plots or performance styles, instead consisting of social nuances, spontaneous, and evolving interaction dynamics. Existing methods often selectively sample discrete frames for model training (e.g., sampling 1 or 2 frames per second within each short video)~\cite{24,25,26,27,28,29}, which inherently fragments the interaction process, making it challenging to model the evolution of human relationships in real-world scenarios (\textit{Challenge 2}).

Recent studies have advanced SRR in long videos through temporal modeling with memory mechanisms~\cite{28}, which enable cumulative encoding of relationship dynamics over time. However, their focus on sequential modeling often fails to model the periodic human interactions~\cite{99}, such as recurrent nonverbal cues (e.g., gestures and facial expressions), repetitive dialogue patterns, and consistent prosodic features. While such memory mechanisms effectively aggregate temporal dependencies, they typically lack explicit modeling of these long-term periodic interaction patterns, making it difficult to model the continuous and longitudinal evolution of real-world relationships. For example, as depicted in Fig.~\ref{Fig: 1}(d), when the video is divided into $N$ clips, different interaction behaviours emerge in each clip, leading to varying interpersonal relationship interpretations when analysed in isolation. An initial interaction, such as a nod or greeting, might lead the model to classify the interpersonal relationship as ``\textit{\textbf{Str}anger}'' (Str), while independent analysis of subsequent segments might yield alternating classifications of ``Str'', ``\textit{\textbf{Acq}uaintance}'' (Acq), or ``Fri''. This phenomenon occurs because social relationships emerge from the cumulative effect of repeated interactions~\cite{38}, and the model interprets behaviours at different stages of the interaction cycle as different relationship patterns. Accurately capturing such periodic temporal signals and understanding repetitive behavioural patterns across time and space is crucial for mitigating inference errors and enhancing the robustness of relationship modeling (\textit{Challenge~3}).

Inspired by these insights and challenges, we propose a novel Multimodal Graph-based Framework for Spatio-temporal \textbf{Asy}mmetric Dyadic \textbf{Re}lationship \textbf{C}lassification (AsyReC), which segments an input video into uniformly continuous temporal clips to facilitate fine-grained learning of dyadic interactions. Specifically, for each segmented clip, we introduce a triplet graph structure augmented with a dual attention mechanism to model the asymmetric social relationships between individuals. For global spatio-temporal modeling, we map the temporal index of each clip onto sine and cosine waveforms to encode periodic temporal patterns. The interaction knowledge derived from the triplet graph is then fused with these temporal signals, enabling the model to capture recurring behavioural patterns over time. Our main contributions are as follows:
\begin{itemize}
    \item A triplet graph neural network architecture with a dual attention mechanism is proposed to model the adaptive contributions of multimodal features and interaction cues, effectively modeling asymmetric relational patterns in social interactions (Addresses \textit{Challenge~1});
    \item A clip-level dyadic relationship learning framework is introduced to capture the temporal evolution of dyadic relationships, enabling fine-grained modeling of dynamic social interactions in real-world scenarios (Addresses \textit{Challenge~2});
    \item A novel spatio-temporal modeling method is introduced to encode periodic interaction patterns by projecting time indices onto sine and cosine waveforms, capturing the global evolution of interaction dynamics (Addresses \textit{Challenge~3});
    \item Extensive evaluations on the NoXi~\cite{49} and UDIVA~\cite{48} datasets demonstrate how AsyReC outperforms several baselines, confirming the effectiveness of the framework.
\end{itemize}

\section{RELATED WORK}
In this section, we provide a comprehensive review of existing methods for social relationship recognition (SRR), including approaches based on image data, video data, and multimodal frameworks. In addition, we systematically summarize the limitations identified in these studies and articulate the novelties underlying our work.

\subsection{Image-based SRR}
Image-based SRR has evolved through three phases: feature-based classification, graph reasoning, and high-order structural modeling. Initially, Convolutional Neural Networks (CNNs) were used for visual feature extraction to predict binary relationships~\cite{66,67}. Li et al.~\cite{11} extended this approach with multi-attribute analysis (appearance, geometry, context), but their dual-glance model suffered from attribute redundancy. Wang et al.~\cite{10} addressed feature selection through multi-source face-body attribute fusion, while Tang et al.~\cite{68} enriched the feature representation by incorporating interpersonal similarity and using a confusion loss function to reinforce correct identification and penalize misclassifications. However, these approaches were limited in capturing implicit social associations, leading to the development of graph-based reasoning.

The breakthrough in graph reasoning began with the proposal of the Graph Reasoning Model~\cite{72} for modeling human-object interaction, which was subsequently refined by two key innovations: global-local feature decoupling~\cite{16} improved representation learning and GRU-based temporal updating~\cite{73} improved dynamic relationship tracking. Li et al.~\cite{13} advanced this by constructing social relationship graphs with GCN-GRU integration. However, contextual under-utilization persisted until Li et al.~\cite{74} pioneered the joint modeling of human-human and human-scene interactions through scene-embedded graph nodes. This foundation was strengthened by Qing et al.~\cite{15} through global-local semantic fusion and Sousa et al.~\cite{75} through prior knowledge integration. Tang et al.~\cite{76} then introduced distributed reasoning strategies to hierarchically process primary and secondary interactions.

Recent advances have focused on modeling higher-order relationships in structural graph networks. Tang et al.~\cite{18} developed advanced graph architectures for individual, dyadic, and group interactions. Guo et al.~\cite{19} and Yu et al.~\cite{20} proposed triadic constraint modeling and multi-level attention mechanisms for complex interaction analysis, respectively. In addition, Tang et al.~\cite{70} introduced graph-based interactive knowledge distillation for multi-stage class-incremental learning. However, these methods rely on static images, which limit their ability to capture the dynamic evolution of human relationships over time. To address this, we propose a clip-level relationship recognition framework to effectively model and exploit the spatio-temporal dynamics of human interactions.

\subsection{Video-based SRR}
In recent years, video-based SRR has emerged as a significant area of research. The seminal work of Lv et al.~\cite{77} established SRR as a video classification task and laid the foundation for subsequent developments. Building on this, Dai et al.~\cite{78} proposed a two-step model that integrates spatio-temporal feature extraction with object semantics to improve relationship recognition. Similarly, Yan et al.~\cite{80} advanced temporal modeling by tracking appearance timelines and constructing dynamic knowledge graphs to capture longitudinal behavioural patterns. In addition, Lv et al.~\cite{81} proposed a sequence recurrent network that incorporates global-local attention mechanisms to prioritize relational discriminative information. Despite their contributions, these approaches mainly use visual signals directly for SRR, and they were limited in modeling the spatio-temporal interaction dynamics that underlie social behaviour.

Recent research has shifted toward graph-structured relationship reasoning frameworks. Liu et al.~\cite{21} proposed a tripartite graph model that links people and contextual objects using Pyramid Graph Convolutional Networks (PGCN) to model temporal dependencies. Teng et al.~\cite{82} developed a character relation reasoning graph to model the dynamics of relationship propagation, while Yu et al.~\cite{23} integrated message-passing mechanisms with spatio-temporal features for pairwise prediction. A key limitation of these methods is their assumption of relationship symmetry, which overlooks the directional nature of social interactions~\cite{33,34}. Our work addresses this gap by modeling asymmetric interaction patterns that capture directional dependencies and variability in real-world relationships.

\subsection{Multimodal-based SRR}
Recent advances have explored multimodal fusion strategies for predicting social relationships by correlating visual-linguistic cues~\cite{24,83}. Liu et al.~\cite{27} pioneered this direction through a multimodal framework that learns global action scene representations and achieves refined relationship extraction through their Multi-Conv Attention module. Later, Wu et al.~\cite{26} proposed a temporal graph aggregation framework that unifies visual, textual, and auditory cues by constructing frame-level subgraphs and aggregating them into a video-level social graph. Teng et al.~\cite{25} developed a pre-training framework that captures spatiotemporal interactions between visual instances and aligns visual and speech features using cross-modal attention, but it did not address long-term relationship evolution. Wang et al.~\cite{28} addressed this limitation with a cumulative transformer framework enhanced by memory mechanisms for long video analysis. 

In addition, Li et al.~\cite{06} investigated two-person interactions by leveraging skeletal data and GNNs, providing insights into the spatio-temporal dynamics of interpersonal exchanges. Saeid et al.~\cite{57} employed a multi-camera setup to capture individual interactions from different angles to comprehensively analyse the nuances of social dynamics. Li et al.~\cite{71} used large language models for zero-shot relationship learning via visual signals and social narratives.

Despite these innovations, existing approaches have not focused on periodic interaction patterns, which are critical for modeling long-term relationship dynamics. To address this, we propose a novel temporal signal mapping method to model periodic interactions, providing a more comprehensive understanding of relationship dynamics.

\section{Problem Formulation}
Building on the previous SRR methods~\cite{13}, we formalize AsyReC as a classification task that aims to estimate the conditional probability distribution of a set of social relationship labels. Specifically, given synchronized dual-view videos capturing two interactants (individuals) $i$ and $j$, we derive a probabilistic mapping using multimodal signals: visual features $\boldsymbol{v}_{i}, \boldsymbol{v}_{j}$, bounding boxes $\boldsymbol{b}_i, \boldsymbol{b}_j$, audio features $\boldsymbol{a}_i, \boldsymbol{a}_j$, linguistic cues $\boldsymbol{l}_i, \boldsymbol{l}_j$, and asymmetric relationship labels $\{\boldsymbol{x}_{i\to j}, \boldsymbol{x}_{j\to i}\}$ for $\forall i,j \in \{1,2,\dots,N\}$ with $i \neq j$, where $N$ is the total number of single-person video pairs. AsyReC optimizes two probability functions independently: 
\begin{equation}
    \label{Eq: 01}
    x_i^{*} = \underset{x_i}{\mathrm{argmax}} \ P(\{\boldsymbol{x}_{i\to j}|\boldsymbol{\Theta}_i,\boldsymbol{\Theta}_j,\mathcal{T}),
\end{equation}
\begin{equation}
    \label{Eq: 02}
    x_j^{*} = \underset{x_j}{\mathrm{argmax}} \ P(\{\boldsymbol{x}_{j\to i}|\boldsymbol{\Theta}_i,\boldsymbol{\Theta}_j,\mathcal{T}),
\end{equation}
where $\boldsymbol{\Theta}_k = (\boldsymbol{v}_k,\boldsymbol{b}_k,\boldsymbol{a}_k,\boldsymbol{l}_k)$ represents the multimodal evidence for person $k \in \{i,j\}$, $\mathcal{T}$ is the time index of the clip, and $x_k^{*}$ is the optimal predictive probability distribution of the relationship from person $k$'s perspective.

\begin{figure*}[t!]
	\setlength{\abovecaptionskip}{0.cm}
	\setlength{\belowcaptionskip}{-0.cm}
	\centering
	\includegraphics[width=5.5in]{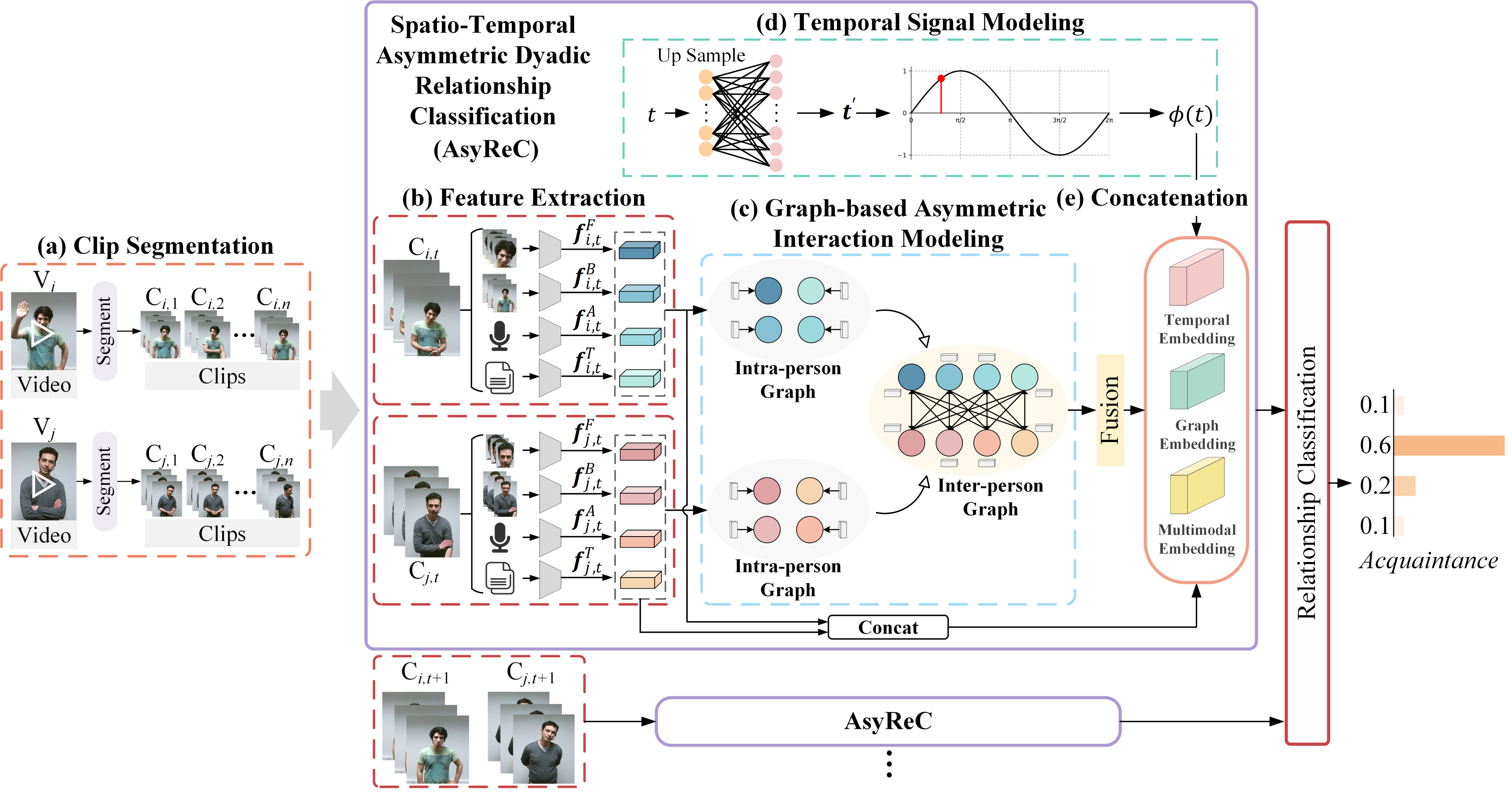}
	\caption{The overall framework of AsyReC. First, (a) a pair of videos is segmented into $n$ clips. (b) Each pair of clips is then processed to extract face, body, audio, and text features using dedicated encoders. (c) These features are structured into graph networks to model asymmetric interactions. (d) Simultaneously, the temporal signals are upsampled into high-dimensional embeddings, followed by sin/cos wave mapping. Finally, (e) the temporal embeddings, multimodal feature embeddings, and graph-inferred knowledge are concatenated for relationship classification. The screenshots are from the NoXi database collected for the ARIA-VALUSPA project~\cite{49}. \label{Fig: 2}}
    \vspace{-1.0em}
\end{figure*}

\section{Method}
This section introduces the proposed AsyReC framework. The pipeline is shown in Fig. \ref{Fig: 2}, including clip segmentation, multimodal feature extraction, graph-based asymmetric interaction modeling, temporal signal modeling, and relationship classification. Input videos are first segmented into $n$ synchronized clip pairs $\{C_{i,t}, C_{j,t}\}$, where modality-specific encoders extract face, body, audio, and text features. These features are then encoded into two intra-person graphs ($G_{intra}^I$, $G_{intra}^J$) to isolate individual behaviours, such as $I$'s frequent smiling versus $J$'s crossed-arm posture. The graphs undergo node attention to enhance discriminative cues (e.g., intensifying $I$'s eyebrow raises while suppressing $J$'s occasional head nods), then merge into one inter-person graph ($G_{inter}$) where edge attention models asymmetric dependencies (e.g., how $I$'s accelerated speech rate reduces $J$'s gesture frequency). Meanwhile, temporal indices are encoded via sinusoidal waveforms to track interaction evolution and periodic interaction dynamics. Finally, fused representations combining raw features, graph-inferred knowledge, and temporal dynamics enable robust classification, effectively modeling spatio-temporal perceptual asymmetry.

\subsection{Clip Segmentation}
Given a pair of videos $V_i$ and $V_j$ (each containing one person and of duration $T$), we segment both into $n$ non-overlapping clips of equal length $\Delta \tau = T/n$. The $t$-th clip $C_{k,t}$ from $V_k$, where $t \in \{1, \ldots, n\}$ and $k \in \left \{ i,j \right \} $, spans an interval ensuring sequential alignment so that each clip starts precisely where the previous one ends (e.g., Clip 1: $[0, \Delta \tau]$, Clip 2: $[\Delta \tau, 2\Delta \tau]$,  $\dots$, Clip $t$: $[(t-1)\Delta \tau, t\Delta \tau]$). 
The audio streams and dialog transcripts are then extracted using the identical time windows $[(t-1)\Delta \tau, t\Delta \tau]$. This generates $n$ aligned multimodal units, each consisting of paired clips $(C_{i,t}, C_{j,t})$, their corresponding synchronized audio, and matching dialog text segments.

\subsection{Feature Extraction}
The AsyReC framework extracts temporally synchronized multimodal features from segmented clips of duration $\Delta\tau$ using modality-specific encoders. For each pair of clips $(C_{i,t}, C_{j,t})$, face features $\boldsymbol{f}_{k,t}^F \in \mathbb{R}^d$ and body features $\boldsymbol{f}_{k,t}^B \in \mathbb{R}^d$ are extracted using 3D spatio-temporal visual encoders~\cite{50}. Audio features $\boldsymbol{f}_{k,t}^A \in \mathbb{R}^d$ are encoded via a pre-trained spectral-temporal transformer~\cite{52}, while textual features $\boldsymbol{f}_{k,t}^T \in \mathbb{R}^d$ are encoded using a pre-trained language model~\cite{53}, with strict alignment to the $\Delta\tau$ temporal windows. The framework ensures congruent temporal granularity and identical dimensionality $\{\boldsymbol{f}_{k,t}^{F,B,A,T}\} \in \mathbb{R}^d, k \in \{i,j\}$ across modalities, thereby facilitating cross-modal alignment. 

\subsection{Graph-based Asymmetric Interaction Modeling}
\begin{figure}[t!]
    \setlength{\abovecaptionskip}{0.cm}
    \setlength{\belowcaptionskip}{-0.cm}
    \centering
    \includegraphics[width=0.8\linewidth]{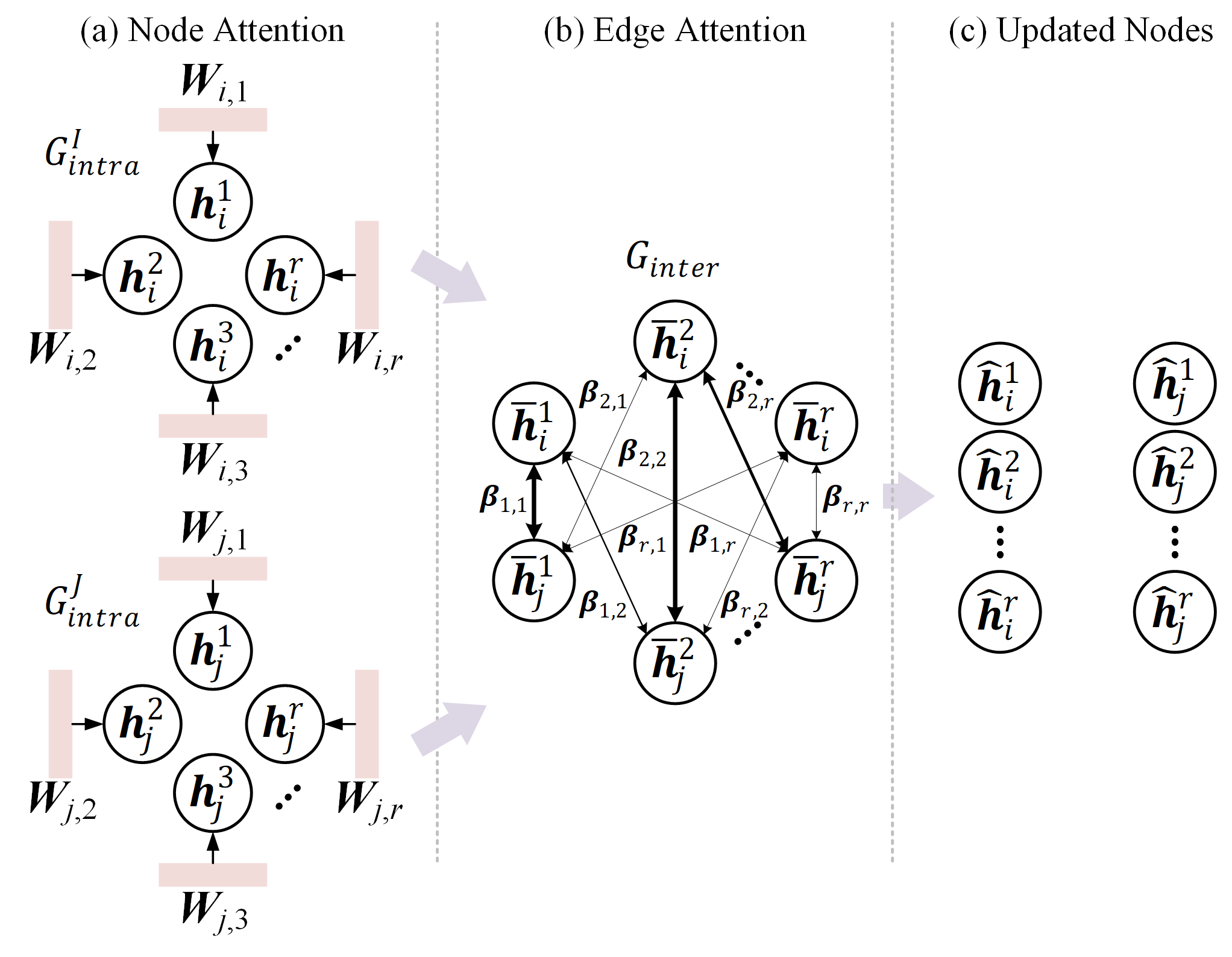}
    \vspace{-0.5em}
    \caption{Node-Edge Attention Graph Network (NE-AGN). The model sequentially computes (a) node attention, (b) edge attention, and (c) updated node representations. \label{Fig: 3}}
    \vspace{-1.0em}
\end{figure}

Dyadic relationships often exhibit perceptual asymmetry, where individuals hold different interpretations of their interactions (e.g., $I$ labels $J$ as a ``\textit{friend}'', while $J$ categorizes $I$ as an ``\textit{acquaintance}''). Traditional graph neural networks (GNNs), such as GCN~\cite{84}, GAT~\cite{85}, and GGNN~\cite{86}, inadequately model such asymmetries due to their uniform treatment of multimodal features across individuals, which obscures distinct behavioural patterns. While GAT introduces node-level attention, it fails to capture dual asymmetric representations or cross-modal dependencies (e.g., how $I$'s facial expressions modulate $J$'s vocal responses). To address these limitations, we propose a Node-Edge Attention Graph Network (NE-AGN) (Fig. \ref{Fig: 3}), which employs a triple graph architecture ($G_{intra}^I$, $G_{intra}^J$, $G_{inter}$) to hierarchically resolve perceptual asymmetry. The model first isolates individual-specific modalities (e.g., prioritizing $I$'s facial cues and $J$'s speech prosody) through intra-person node attention, then models asymmetric cross-entity interactions (e.g., $I$'s dominant gestures influencing $J$'s responsive posture) via inter-person edge attention. This tiered approach ensures that individual uniqueness and dyadic relationship dynamics are jointly encoded without mutual interference, enabling robust modeling of both individual behaviour and asymmetric relationship patterns.

Specifically, for $r$ modalities at time index $t$, each feature $\boldsymbol{f}_{k,t}^{r}$ is encoded as a node in a graph, where the node set $\boldsymbol{h}_{k}=[\boldsymbol{h}_{k}^{1},\ldots,\boldsymbol{h}_{k}^{r}]$ satisfies $\boldsymbol{h}_{k}^{r}=\boldsymbol{f}_{k,t}^{r}$ with $k \in \{i,j\}$. Node attention adaptively scales modalities using a learnable tensor $\boldsymbol{\Omega}_{k}^{r}$ and is followed by residual updates:  
\begin{equation}
    \label{Eq: 05}
    \boldsymbol{W}_{k,r} = softmax(\text{LeakyReLU}(\boldsymbol{h}_{k}^{r} \times \boldsymbol{\Omega}_{k}^{r})),
\end{equation}  
\begin{equation}
    \label{Eq: 06}
    \overline{\boldsymbol{h}}_{k}^{r} = \boldsymbol{h}_{k}^{r} + \boldsymbol{h}_{k}^{r} \cdot \boldsymbol{W}_{k,r}.
\end{equation}  
This node attention weighting helps the model focus on each person's unique multimodal cues (e.g., prioritizing gestures from expressive individuals) while suppressing extraneous signals. To model inter-person asymmetric interactions, the updated nodes are fully connected via the adjacency matrix:
\setlength{\jot}{3pt}
\begin{equation}
    \label{Eq: 07}
    \boldsymbol{A}_{i,j} = 
    \begin{cases}
        1, & i \neq j \\
        0, & i = j
    \end{cases}
\end{equation}
which ensures connections only exist between individuals $i$ and $j$, with no intra-individual connections. Edge attention is then computed cross-entity weights $\boldsymbol{\beta}_{u,v}$ via concatenated features:  
\begin{equation}
    \label{Eq: 08}
    \boldsymbol{\beta}_{u,v} = \text{softmax}(\boldsymbol{\Phi}_{u,v} \times [\boldsymbol{\omega}\overline{\boldsymbol{h}}_{i}^{u} \oplus \boldsymbol{\omega}\overline{\boldsymbol{h}}_{j}^{v}]), \quad u \neq v, \ u,v \in [1, r],
\end{equation}  
where $\boldsymbol{\Phi}_{u,v}$ is a learnable matrix and $\oplus$ denotes concatenation. This approach enables the model to prioritize key multimodal interactions between individuals, which are essential for detecting asymmetries (e.g., a change in $J$'s expression in response to $I$'s increased vocal pitch, altering their relational dynamics).

Finally, node embeddings are updated as follows: 
\begin{equation}
    \label{Eq: 09}
    \widehat{\boldsymbol{h}}_{k}^{v} = \text{ReLU}\left( \sum_{u=1}^{r} \boldsymbol{\beta}_{k}^{u,v}\boldsymbol{a}_{u,v}\overline{\boldsymbol{h}}_{k}^{u} \right),
\end{equation}  
where $\widehat{\boldsymbol{h}}_{k} = [\widehat{\boldsymbol{h}}_{k}^{1}, \ldots, \widehat{\boldsymbol{h}}_{k}^{r}]$. The proposed triple graph structure effectively preserves individual distinctions in $G_{intra}$ while modeling dyadic interactions in $G_{inter}$. $\widehat{\boldsymbol{h}}_{k}$ are then merged through average pooling to form a comprehensive representation of graph inference knowledge.

\subsection{Temporal Signal Modeling}  
\begin{figure*}[t!]
    \vspace{-1.0em}
    \setlength{\abovecaptionskip}{0.cm}
    \setlength{\belowcaptionskip}{-0.cm}
    \centering
    \includegraphics[width=0.7\linewidth]{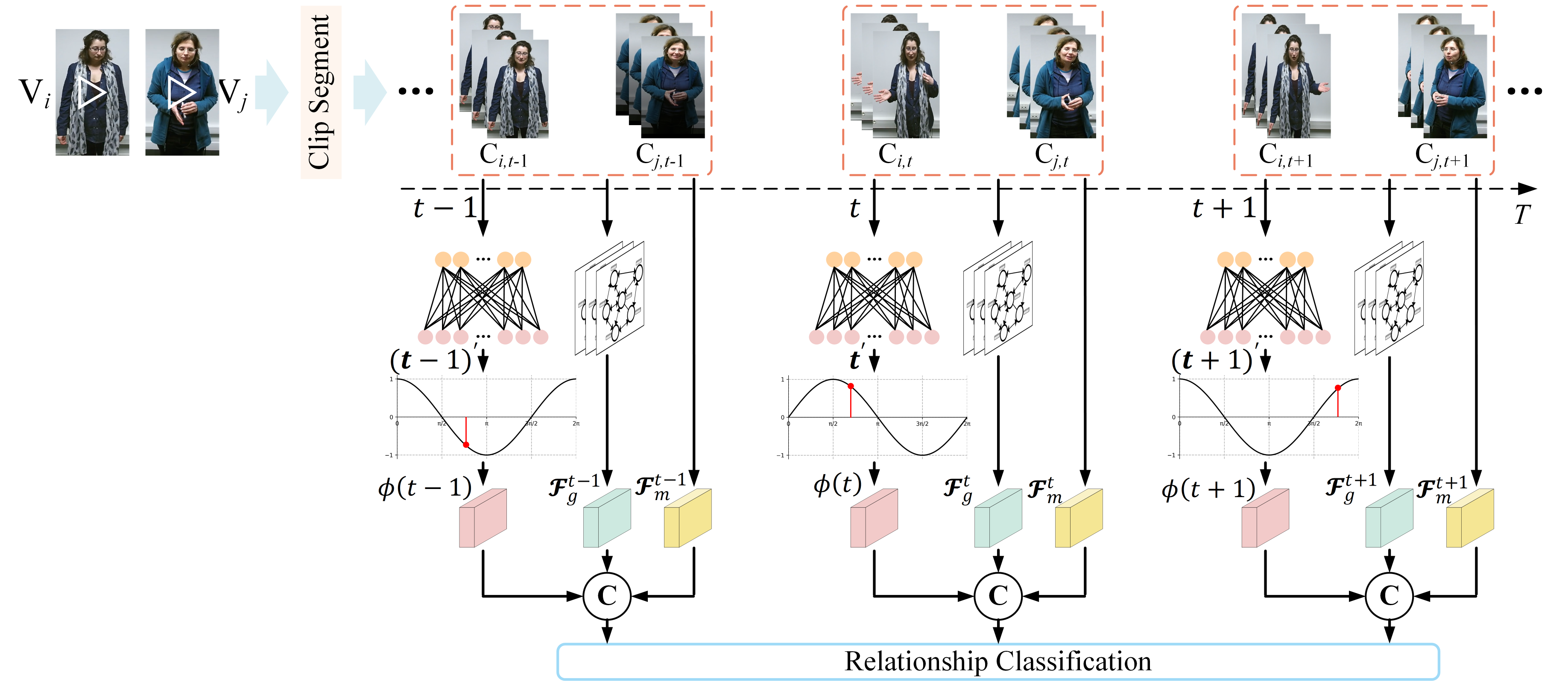}
    \caption{Temporal Signal Modeling Framework. Given an input video pair, it is segmented into $T$ clip pairs, each processed through multimodal feature encoding, graph-based interaction inference, and temporal signal encoding. Encoded temporal signals, graph-inferred knowledge, and multimodal features are fused for relationship classification. Classification layer weights are shared across clips, enabling automatic learning of periodic dependencies via temporal embedding. The screenshots are from the NoXi database collected for the ARIA-VALUSPA project~\cite{49}. \label{Fig: 4}}
    \vspace{-1.0em}
\end{figure*}

Social interactions often exhibit periodic patterns, such as question-answer cycles, synchronized facial expressions (e.g., shared laughter), rhythmic body movements (e.g., nodding), and prosodic variations (e.g., pitch modulation). Sine and cosine waves, as inherently periodic functions with consistent intervals~\cite{65}, are well-suited for modeling such temporal dynamics due to three key properties~\cite{87}: (1) smooth differentiability ensures stable gradient propagation for learning subtle temporal variations; (2) frequency invariance under temporal shifts preserves spectral integrity in recurrent cycles; and (3) phase-magnitude decoupling explicitly models asynchronous coordination. Leveraging these properties, we propose a temporal encoding framework (Fig. \ref{Fig: 4}) that integrates temporal upsampling with periodic-aware trigonometric mapping. Raw clip indices are upsampled via fully connected linear networks~\cite{63}, which project the temporal signal $t$ into a high-dimensional latent space aligned with graph embeddings. These representations are then mapped onto $2\pi$-periodic sine or cosine waves, enabling the framework to capture recurrent interaction dynamics with smooth, frequency-specific periodicity.

Specifically, raw clip indices $t \in \{1, 2, \dots, T\}$ are upsampled into a high-dimensional tensor $\boldsymbol{t}^{'}$:  
\begin{equation}  
	\label{Eq: 10}  
	\boldsymbol{t}^{'} = \boldsymbol{W}_{t} \cdot t + \boldsymbol{b}_{t},  
\end{equation}  
where $\boldsymbol{W}_{t} \in \mathbb{R}^{d \times 1}$ projects $t$ into a $d$-dimensional latent space. Upsampled temporal tensors $\boldsymbol{t}^{'}$ are then mapped onto $2\pi$-periodic sine waves or cosine waves:
\begin{equation}
    \label{Eq: 11}
    \phi(t) = \begin{cases}
        \sin\left(\frac{2\pi \cdot \boldsymbol{t}^{'}}{\max(\boldsymbol{T}^{'},\varepsilon)}\right), & \text{if } t \text{ is even},\\
        \cos\left(\frac{2\pi \cdot \boldsymbol{t}^{'}}{\max(\boldsymbol{T}^{'},\varepsilon)}\right), & \text{if } t \text{ is odd},
    \end{cases}
\end{equation}
where $\phi(t)$ is the encoded temporal signal, $\boldsymbol{T}^{'}$ is the maximum upsampled index, and $\varepsilon \ll 1$ prevents division by zero. The $\pi$-scaled periodic basis ensures full oscillation cycles across the time axis, effectively modeling local and global periodic interpersonal interaction patterns.

\subsection{Concatenation and Classification}
Finally, multimodal feature embeddings ($\boldsymbol{\mathcal{F}}_{m}$), graph-inferred knowledge ($\boldsymbol{\mathcal{F}}_{g}$), and encoded temporal signals ($\phi(t)$) are concatenated into a unified spatio-temporal embedding. This embedding is passed through a fully connected layer (FC) followed by a softmax function to produce a probability distribution over the relationship classes:
\begin{equation}
    \label{Eq: 12}
    R = \arg\max \left \{ softmax\left(\text{FC}(\boldsymbol{\mathcal{F}}_{m}^{t}, \boldsymbol{\mathcal{F}}_{g}^{t}, \phi(t))\right) \right \},
\end{equation}
where $\mathbf{p} = [p_1, p_2, \dots, p_n]$ is the predicted probability distribution over the $n$ relationship classes. The final prediction $R$ is the class with the highest probability.

The framework is trained using the cross-entropy loss, which measures the discrepancy between the predicted probability distribution and the true class labels:
\begin{equation}
    \label{Eq: 13}
    L_{c} = -\sum_{m=1}^{n} \mathbb{I}(y_c = m) \log(p_{m}),  
\end{equation}  
where $\mathbb{I}(y_c = m)$ is an indicator function that equals 1 if the true class label $y_c$ matches the class index $m$, and 0 otherwise. $p_{m}$ denotes the predicted probability for class $m$.

\section{Evaluation Setup}  
\subsection{Datasets}  
To evaluate our method, we used two public datasets: NoXi \cite{49} and UDIVA \cite{48}. The NoXi dataset comprises 168 videos (84 dyadic interactions), totaling over 51 hours of recordings, with an average duration of 18 minutes per video. It includes four relationship categories: \textit{\textbf{Str}angers} (Str.), \textit{\textbf{Acq}uaintances} (Acq.), \textit{\textbf{Fri}ends} (Fri.), and \textit{\textbf{V}ery \textbf{g}ood \textbf{f}riends} (Vgf.). In contrast, UDIVA dataset contains 1,160 videos (580 dyadic interactions), totaling nearly 130 hours of recordings, with an average length of 6.5 minutes per video, and two relationship categories: \textit{\textbf{Kno}wn} (Kno.) and \textit{\textbf{Unk}nown} (Unk.). Raw videos were processed into clip segmentations following the procedure in Sec. IV-A, with the resulting distribution shown in Table \ref{Table: 1}. Notably, NoXi relationships are bidirectional, varying based on each participant’s own perspective of the interaction, while UDIVA contains unidirectional relationships. These datasets help validate the generalizability of the proposed method.

\begin{table}[h]\footnotesize
	\caption{Data Distribution for NoXi and UDIVA.}
	\vspace{-0.5em}
	\centering
	\def\temptablewidth{0.2\textwidth}
	\tabcolsep=0.05cm
	\renewcommand\arraystretch{1.5}
	\begin{threeparttable}
			\begin{tabular}{c|cccc|cccc|cc}
				\hline
				& \multicolumn{4}{c|}{NoXi-I}                                                                 & \multicolumn{4}{c|}{NoXi-J}                                                                  & \multicolumn{2}{c}{UDIVA}            \\ \cline{2-11} 
				& \multicolumn{1}{c|}{Str.}  & \multicolumn{1}{c|}{Acq.}  & \multicolumn{1}{c|}{Fri.} & Vgf.  & \multicolumn{1}{c|}{Str.}  & \multicolumn{1}{c|}{Acq.}  & \multicolumn{1}{c|}{Fri.}  & Vgf.  & \multicolumn{1}{c|}{Kno.}   & Unk.   \\ \hline
				videos & \multicolumn{1}{c|}{36}    & \multicolumn{1}{c|}{30}    & \multicolumn{1}{c|}{6}    & 12    & \multicolumn{1}{c|}{37}    & \multicolumn{1}{c|}{26}    & \multicolumn{1}{c|}{12}    & 9     & \multicolumn{1}{c|}{512}    & 648    \\ \hline
				clips  & \multicolumn{1}{c|}{3,792} & \multicolumn{1}{c|}{3,320} & \multicolumn{1}{c|}{643}  & 1,392 & \multicolumn{1}{c|}{3,642} & \multicolumn{1}{c|}{3,045} & \multicolumn{1}{c|}{1,144} & 1,098 & \multicolumn{1}{c|}{20,024} & 24,526 \\ \hline
			\end{tabular}
			\footnotesize \textit{Note: In the NoXi dataset, NoXi-I and NoXi-J denote the relationships between the two people as perceived from their individual perspective. All data is presented in terms of quantities; for example, there are 36 videos for Str. and 3,792 corresponding clips in NoXi-I.}
		\end{threeparttable}
	\label{Table: 1}
    \vspace{-1.5em}
\end{table}

\subsection{Baseline Models}
We considered the following video-based SRR baseline models to evaluate the proposed AsyReC framework:
\begin{itemize}
    \item \textbf{PGCN}~\cite{21} introduces a pyramid graph convolutional network that uses visual signals to construct triple graphs (intra-person, inter-person, and person-object graphs) for modeling dynamic interactions and scene contexts. 
    \item \textbf{LIReC}~\cite{24} proposes a multimodal framework for jointly predicting individual interactions and relationships in movies by integrating visual and textual features. 
    \item \textbf{Cumulative Transformer (CT)}~\cite{28} introduces a cumulative memory mechanism within a Transformer architecture, extending the temporal receptive field to aggregate historical information for long-term relationship inference.
\end{itemize}

\subsection{Implementation Details}
\textbf{Visual Signal Processing:} Input videos were downsampled from 25 fps to 5 fps and divided into 10-second clips to balance the reduction in temporal resolution with computational efficiency while preserving essential interaction patterns, following the approach of~\cite{96}. Body and face regions were detected using YOLO v8~\cite{41} and MTCNN~\cite{40}, respectively, similarly to~\cite{28}. Both regions were resized to $224 \times 224$ pixels and augmented with random horizontal flipping and normalization, in a manner similar to~\cite{21,24,28}. The spatio-temporal features were then extracted using two separate 3D ResNet-18~\cite{50} models pre-trained on Kinetics-400 similarly to~\cite{97}.

\textbf{Audio Signal Processing:} Raw audio streams were demultiplexed from synchronized video using FFmpeg~\cite{42}. Acoustic representations were extracted using librosa~\cite{43} to capture prosodic, harmonic, and timbral features in a manner similar to~\cite{05}. These features were structured into NumPy arrays, encoded into audio embeddings using an Audio Spectrogram Transformer~\cite{52}, and temporally aligned with corresponding video clips for a synchronized multimodal process, following a similar approach as~\cite{05}.

\textbf{Textual Signal Processing:} The temporally aligned audio transcripts were first generated using Whisper~\cite{49} for speech-to-text conversion. Text embeddings were then extracted using BERT~\cite{53} to facilitate further textual analysis and processing, as suggested by~\cite{28}.

\textbf{Training and Testing Strategy:} As shown in Table \ref{Table: 1}, there is a pronounced data imbalance within the NoXi and UDIVA datasets. To effectively evaluate the proposed model, we implemented a $K$-fold cross-validation strategy \cite{55} with $K=3$. This methodological approach is designed to mitigate the effects of data imbalance, thereby allowing for a fair assessment of the effectiveness of the proposed model.

\textbf{Experimental Setup:} All experiments were implemented using the PyTorch framework~\cite{54} and run on a high-performance computing server equipped with an Intel(R) Xeon(R) Gold 5317 CPU @ 3.00GHz and an NVIDIA A800 GPU (128 GB memory). The hyperparameters were configured as follows: a batch size of 64, a learning rate of 0.0001 with an L2 regularization weight decay of 0.0005, and a maximum training duration of 200 epochs. To avoid overfitting and to optimize training efficiency, early stopping was used, which stops the training process if no improvement in validation performance was observed for a predefined number of consecutive epochs.

\subsection{Evaluation Metrics}
Given the class imbalance in the NoXi and UDIVA datasets, we adopted three metrics to mitigate evaluation bias caused by skewed category distributions, as suggested in~\cite{21,24,28}. The class-specific recall for category $i$ is defined as:
\begin{equation}
    \mathcal{R}_i = \frac{TP_i}{TP_i + FN_i},
\end{equation}
where $TP_i$ and $FN_i$ denote true positives and false negatives for class $i$. We used the Unweighted Average Recall (UAR)~\cite{56}, which aggregates the performance uniformly over all classes. This was adopted to mitigate the influence of class imbalance and to ensure equitable evaluation across categories with uneven sample distributions.

In addition, to evaluate the importance of each component in the framework, we used a masking approach to evaluate the predictions. The fidelity metric ($\Delta F$)~\cite{59,62} quantifies the average difference between the original model $\mathcal{O}(\mathbf{x}_n)$ and its masked variants $\mathcal{P}(\mathbf{x}_n)$ over all observations:
\begin{equation}
    \Delta F = \frac{1}{K}\sum_{n=1}^{K} \| \mathcal{O} (\mathbf{x}_n) - \mathcal{P} (\mathbf{x}_n) \|_1,
\end{equation}
where $K$ is the number of cross-validation folds. A higher $\Delta F$ value indicates a greater importance of the masked component and is useful to evaluate which variables contribute more to cross-modal interactions and periodic temporal signal modeling in the AsyReC framework.

\section{Experiments and Results}
\subsection{Comparative Analyses}
\subsubsection{Performance Comparison}
To evaluate the effectiveness of AsyReC, we performed comparative analyses with the baseline models and the proposed framework. For comparison purposes, category-specific recall scores were reported as means, while the overall UAR was computed as a mean and standard deviation. The recall provides a clear accuracy per class, facilitating direct comparison across relationship categories. In contrast, the mean and standard deviation of the UAR reflect the overall stability of the model's performance, highlighting its ability to model class-imbalanced scenarios.

\begin{table*}[t!]\footnotesize
	\caption{Comparison of experimental results with different baselines on NoXi and UDIVA.}
	\vspace{-0.5em}
	\centering
	\def\temptablewidth{0.2\textwidth}
	\tabcolsep=0.15cm
	\renewcommand\arraystretch{1.5}
	\begin{threeparttable}
            \begin{tabular}{c|ccccccccccc|ccc}
            \hline
                & \multicolumn{11}{c|}{NoXi} & \multicolumn{3}{c}{}                                      \\ \cline{2-12}
                & \multicolumn{5}{c|}{NoXi-I} & \multicolumn{5}{c|}{NoXi-J}& NoXi(mean)                                & \multicolumn{3}{c}{\multirow{-2}{*}{UDIVA}}                                                                                  \\ \cline{2-15} 
                & Str.                         & Acq.                                  & Fri.      & \multicolumn{1}{c|}{Vgf.}     & \multicolumn{1}{c|}{UAR}            & Str.                         & Acq.        & Fri.                         & \multicolumn{1}{c|}{Vgf.}      & \multicolumn{1}{c|}{UAR}              & UAR    & Kno.                         & \multicolumn{1}{c|}{Unk.}         & UAR                                       \\ \hline
                PGCN~\cite{21}            & 21.4                         & 19.1                                  & 0.0                                   & \multicolumn{1}{c|}{64.1}                                  & \multicolumn{1}{c|}{26.2±4.0}                                  & 19.4                         & 21.8                                  & \textbf{50.7}                & \multicolumn{1}{c|}{17.9}                                  & \multicolumn{1}{c|}{27.5±7.6}                                  & \cellcolor[HTML]{EFEFEF}26.8±6.6          & \textbf{72.3}                & \multicolumn{1}{c|}{29.9}                         & \cellcolor[HTML]{EFEFEF}51.1±5.2          \\
                LIReC~\cite{24}           & 51.6                         & 45.4                                  & 19.3                                  & \multicolumn{1}{c|}{33.5}                                  & \multicolumn{1}{c|}{37.4±3.1}                                  & \textbf{70.6}                & 45.4                                  & 19.0                         & \multicolumn{1}{c|}{10.4}                                  & \multicolumn{1}{c|}{36.4±13.9}                                 & \cellcolor[HTML]{EFEFEF}36.9±9.9          & 37.7                         & \multicolumn{1}{c|}{\textbf{69.0}}                & \cellcolor[HTML]{EFEFEF}53.3±6.7          \\
                CT~\cite{28}              & \textbf{61.6}                & 57.2                                  & 33.2                                  & \multicolumn{1}{c|}{8.6}                                   & \multicolumn{1}{c|}{40.2±16.3}                                 & 47.7                         & 57.6                                  & 30.0                         & \multicolumn{1}{c|}{25.1}                                  & \multicolumn{1}{c|}{40.1±11.1}                                 & \cellcolor[HTML]{EFEFEF}40.1±13.9         & 70.3                         & \multicolumn{1}{c|}{35.7}                         & \cellcolor[HTML]{EFEFEF}53.0±3.7          \\
                \textbf{AsyReC (ours)} & \cellcolor[HTML]{EFEFEF}49.6 & \cellcolor[HTML]{EFEFEF}\textbf{61.9} & \cellcolor[HTML]{EFEFEF}\textbf{38.4} & \multicolumn{1}{c|}{\cellcolor[HTML]{EFEFEF}\textbf{52.8}} & \multicolumn{1}{c|}{\cellcolor[HTML]{EFEFEF}\textbf{50.7±3.6}} & \cellcolor[HTML]{EFEFEF}47.2 & \cellcolor[HTML]{EFEFEF}\textbf{59.7} & \cellcolor[HTML]{EFEFEF}40.5 & \multicolumn{1}{c|}{\cellcolor[HTML]{EFEFEF}\textbf{35.4}} & \multicolumn{1}{c|}{\cellcolor[HTML]{EFEFEF}\textbf{45.7±4.3}} & \cellcolor[HTML]{EFEFEF}\textbf{48.2±6.5} & \cellcolor[HTML]{EFEFEF}65.5 & \multicolumn{1}{c|}{\cellcolor[HTML]{EFEFEF}53.3} & \cellcolor[HTML]{EFEFEF}\textbf{59.4±6.8} \\ \hline
                \end{tabular}
		\footnotesize \textit{Note: We use UAR (in \%) for overall performance and recall (in \%) for each class.}
	\end{threeparttable}
	\label{Table: 2}
	\vspace{-1.0em}
\end{table*}

The experimental results in Table~\ref{Table: 2} demonstrate AsyReC's superior performance in modeling dyadic interactions compared to PGCN, LIReC, and CT. Our framework achieves state-of-the-art results, with mean UARs of 48.2\% on NoXi and 59.4\% on UDIVA, representing absolute improvements of 6.1-21.4\% over these baseline methods. Among them, PGCN's triple graph structure has significant limitations in capturing asymmetric relationship dynamics, as evidenced by its 0.0\% recall for \textit{Fri} relationships on NoXi-I. In contrast, AsyReC's node-edge dual attention mechanism enables dynamic differentiation of behavioural roles and focuses on asymmetric interaction patterns, leading to significant performance gains. Although PGCN achieves higher recall than AsyReC in certain categories (e.g., Vgf on NoXi-I, Fri on NoXi-J, and Kno on UDIVA), this comes at the cost of severely compromised recognition in other classes, whereas AsyReC provides a more balanced and robust classification across all classes.

AsyReC's multimodal integration also outperforms LIReC's feature fusion approach in most metrics. While LIReC relies solely on body and text features, resulting in considerable performance variability across relationship categories (e.g., 70.6\% for \textit{Str} vs. 10.4\% for \textit{Vgf} on NoXi-J), AsyReC incorporates facial expressions, body gestures, speech prosody, and text semantics. This comprehensive multimodal approach enables more balanced inferences, improving LIReC's \textit{Vgf} recall on NoXi-J from 10.4\% to 35.4\%. However, LIReC outperforms AsyReC in the NoXi-J \textit{Str} category (70.6\% vs. 47.2\%), likely due to the importance of explicit body and verbal cues in stranger interactions, which LIReC's focused modality set captures in more detail. Similarly, LIReC achieves higher recall in UDIVA's \textit{Unk} category (69.0\% vs. 53.3\%), suggesting that unknown relationships may benefit from LIReC's direct feature fusion without periodic modeling constraints. Nevertheless, AsyReC still achieves the best classification accuracy and maintains balance across all relationship categories.

Compared to CT's cumulative memory approach, AsyReC's periodic interaction modeling demonstrates superior long-term reasoning, as evidenced by a 44.1\% improvement in \textit{Vgf} recall (52.8\% vs. CT's 8.6\% on NoXi-I). The performance gap arises from CT's sequential aggregation of all historical states, which accumulates increasingly noisy signals as iterations progress, ultimately degrading model performance. In contrast, AsyReC models periodic interactive behaviour over time, and by fusing salient interaction patterns rather than accumulating all historical information, AsyReC mitigates the progressive accumulation of confounding signals, resulting in more robust performance. Still, we observe that CT outperforms AsyReC in the NoXi-I \textit{Str} category (61.6\% vs. 49.6\%), potentially suggesting that the stranger relationship category relies less on periodic behaviours and more on interaction history. Nevertheless, AsyReC achieves strong generalization across multiple relationships while avoiding overfitting to the specific patterns of any single relationship category.

\subsubsection{Confusion matrices}
\begin{figure}[h!]
	\setlength{\abovecaptionskip}{0.cm}
	\setlength{\belowcaptionskip}{-0.cm}
	\centering
	\includegraphics[width=3.5in]{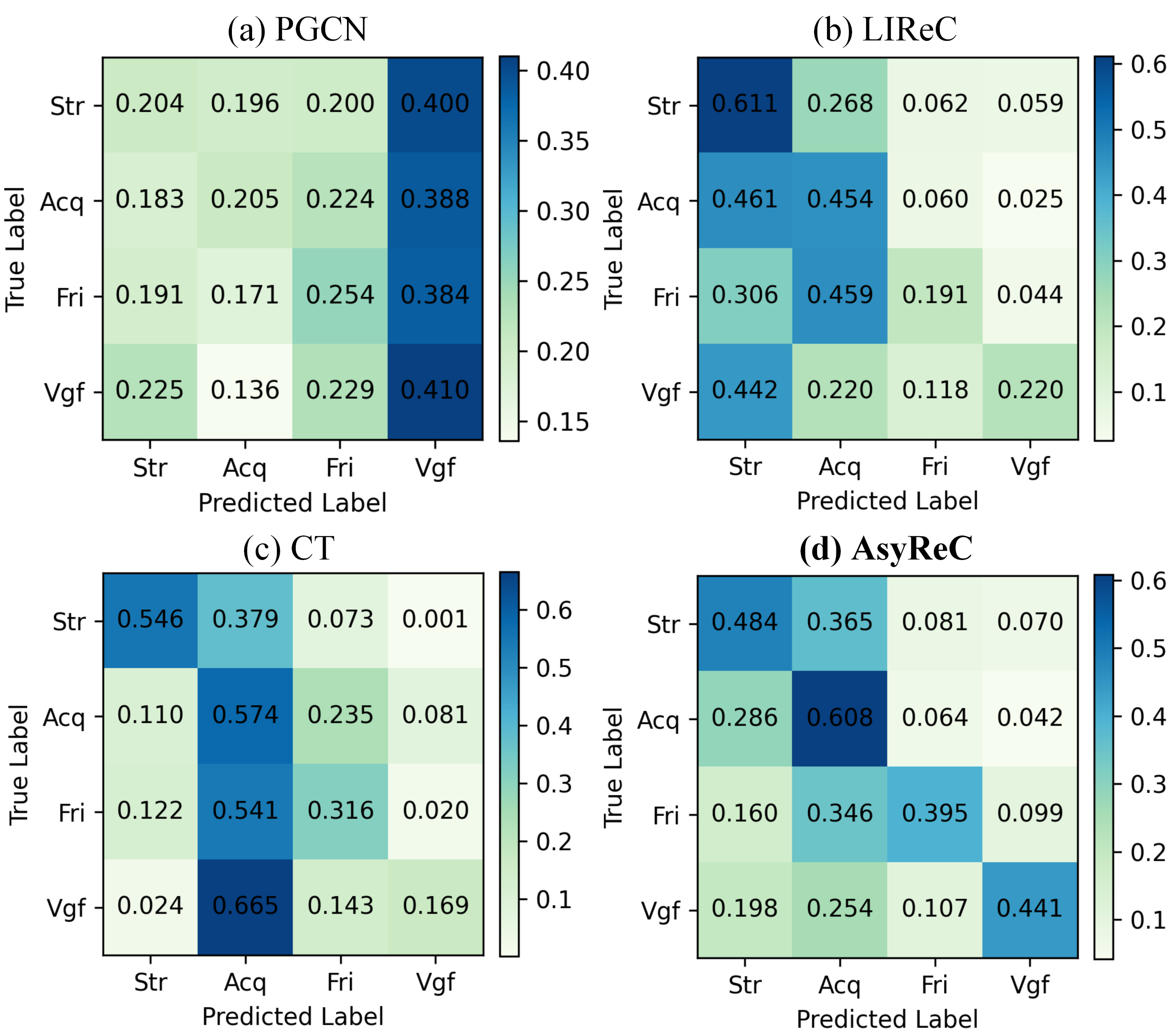}
	\vspace{-1.5em}
	\caption{Confusion matrices for recognition results on NoXi: (a) PGCN, (b) LIReC, (c) CT and \textbf{(d) AsyReC}. \textit{Note: For better visualization, we compute the averages of NoXi-I and NoXi-J}.\label{Fig: 5}}
    \vspace{-1.0em}
\end{figure}

In addition to model accuracies, we also obtained the confusion matrices from the baseline methods and DRR model to gather further insights. The confusion matrices in Fig. \ref{Fig: 5} show clear performance differences between the baseline models and the proposed AsyReC framework. PGCN shows a pronounced bias towards misclassifying relationships as ``very good friends'', with a diagonal value of 0.410, indicating its inability to effectively discriminate between diverse social relationships as it disproportionately aggregates them into the ``Vgf'' category. LIReC, using multimodal fusion, achieves improved performance in the ``Stranger'' category (diagonal value of 0.611), but shows significant weaknesses, correctly identifying only 0.191 of ``Friend'' and 0.220 of ``Vgf'' instances, most of them misclassified as ``Str'' or ``Acquaintance'' (Aqc). CT shows marginal improvement in ``Fri'' recognition (0.316 diagonal value) due to its long-term interaction modeling but performs poorly for ``Vgf'' (0.169), reflecting the accumulation of misclassifications and biases over extended time scales. In contrast, AsyReC demonstrates robust and balanced classification across all categories, achieving significantly higher diagonal values for ``Vgf'' (0.441) and ``Fri'' (0.395), while maintaining strong accuracy for ``Str'' (0.484) and ``Aqc'' (0.608). These results highlight AsyReC's strong performance in discriminating different relationships compared to the baseline methods.

\subsection{Ablation Analysis}
\subsubsection{Ablation of the Entire Framework}
To quantify the contribution of each module, we conduct a comprehensive ablation study with the following configurations:
\begin{itemize}
    \item \textbf{F}: Recognition using facial features only, isolating the impact of facial cues.
    \item \textbf{F+B}: Joint use of facial and body cues to evaluate the complementary role of the body language.
    \item \textbf{F+B+A}: Incorporation of audio features to explore the influence of vocal interactions.
    \item \textbf{F+B+T}: Replacement of audio with textual cues to investigate linguistic cue contributions.
    \item \textbf{F+B+A+T}: Multimodal fusion of face, body, audio, and textual features via concatenation.
    \item \textbf{F+B+A+T+G}: Integration of multimodal signals into the proposed graph network, modeling asymmetric interactions.
    \item \textbf{Full+Dynamic}: Extension of \textbf{F+B+A+T+G} with temporal signal modeling to capture dynamic relationship patterns.
\end{itemize}

The results in Table \ref{Table: 3} show incremental improvements across all datasets. Unimodal face features (F) show limitations, especially in the recognition of intimate relationships (NoXi-J Vgf: 0.5\%). Adding body features (F+B) slightly improves robustness (NoXi-I UAR: 38.0\% vs. 35.3\% for F), but introduces perspective-specific biases. Multimodal integration yields significant gains: audio (F+B+A) improves accuracy via prosodic cues (NoXi-I Fri: 33.2\%, +15.8\% vs. F+B), while text (F+B+T) improves semantic context for NoXi-I Vgf (35.7\% vs. 18.7\% for F+B). However, text integration reduces Vgf recognition in NoXi-J (25.5\% vs. 19.9\% for F+B), revealing the difficulty of modeling asymmetric interaction patterns, i.e., textual signals proved beneficial for one participant's relationship recognition while degrading performance for the other. 

This limitation was effectively addressed by full multimodal integration (F+B+A+T), which achieved overall balanced performance. Graph-based fusion (F+B+A+T+G) further resolved cross-modal asymmetries, significantly improving NoXi-I Vgf (42.7\%, +10.0\% vs. F+B+A+T) and NoXi-J Acq (66.9\%, +14.3\% vs. F+B+A+T). Temporal signal integration (overall) achieves the best performance across all datasets, with pronounced improvements for intimate relationships in NoXi-I (Fri: 38.4\%, Vgf: 52.8\%) and NoXi-J, underscoring the importance of capturing periodic interaction patterns.

\begin{table*}[t!]\footnotesize
	\caption{Experimental results of relationship recognition on the NoXi and UDIVA dataset.}
	\vspace{-0.5em}
	\centering
	\def\temptablewidth{0.2\textwidth}
	\tabcolsep=0.15cm
	\renewcommand\arraystretch{1.5}
	\begin{threeparttable}
            \begin{tabular}{c|ccccccccccc|ccc}
            \hline
            & \multicolumn{11}{c|}{NoXi}
            & \multicolumn{3}{c}{}
            \\ \cline{2-12}
            & \multicolumn{5}{c|}{NoXi-I}
            & \multicolumn{5}{c|}{NoXi-J}
            & NoXi (mean)  & \multicolumn{3}{c}{\multirow{-2}{*}{UDIVA}}                          \\ \cline{2-15} 
                      & Str.      & Acq.                          & Fri.                                   & \multicolumn{1}{c|}{Vgf.}                                   & \multicolumn{1}{c|}{UAR}                                       & Str.                          & Acq.                          & Fri.                                   & \multicolumn{1}{c|}{Vgf.}                                   & \multicolumn{1}{c|}{UAR}                                       & mUAR                                       & Kno.                          & \multicolumn{1}{c|}{Unk.}                          & UAR                                       \\ \hline
            F         & 51.40                         & 58.70                         & 21.20                                  & \multicolumn{1}{c|}{10.00}                                  & \multicolumn{1}{c|}{35.3±18.2}                                 & 53.50                         & 62.70                         & 23.10                                  & \multicolumn{1}{c|}{0.50}                                   & \multicolumn{1}{c|}{35.0±13.9}                                 & \cellcolor[HTML]{EFEFEF}35.2±16.2         & 52.40                         & \multicolumn{1}{c|}{52.00}                         & \cellcolor[HTML]{EFEFEF}52.2±9.1          \\
            F+B       & 53.70                         & 62.10                         & 17.40                                  & \multicolumn{1}{c|}{18.70}                                  & \multicolumn{1}{c|}{38.0±6.7}                                  & 56.70                         & 54.70                         & 20.00                                  & \multicolumn{1}{c|}{19.90}                                  & \multicolumn{1}{c|}{37.8±6.6}                                  & \cellcolor[HTML]{EFEFEF}37.9±6.6          & 58.20                         & \multicolumn{1}{c|}{49.10}                         & \cellcolor[HTML]{EFEFEF}53.6±8.9          \\
            F+B+A     & 55.30                         & 64.80                         & 33.20                                  & \multicolumn{1}{c|}{27.30}                                  & \multicolumn{1}{c|}{45.1±9.2}                                  & 55.70                         & 56.80                         & 25.60                                  & \multicolumn{1}{c|}{27.00}                                  & \multicolumn{1}{c|}{41.3±3.5}                                  & \cellcolor[HTML]{EFEFEF}43.2±8.4          & 63.10                         & \multicolumn{1}{c|}{46.80}                         & \cellcolor[HTML]{EFEFEF}54.9±3.3          \\
            F+B+T     & 46.50                         & \textbf{68.70}                & 30.70                                  & \multicolumn{1}{c|}{35.70}                                  & \multicolumn{1}{c|}{45.4±6.4}                                  & 55.60                         & 55.70                         & 29.10                                  & \multicolumn{1}{c|}{25.50}                                  & \multicolumn{1}{c|}{41.5±3.8}                                  & \cellcolor[HTML]{EFEFEF}43.4±7.0          & 61.40                         & \multicolumn{1}{c|}{47.80}                         & \cellcolor[HTML]{EFEFEF}54.6±3.8          \\
            F+B+A+T   & \textbf{58.00}                & 58.30                         & 36.10                                  & \multicolumn{1}{c|}{32.70}                                  & \multicolumn{1}{c|}{46.3±6.8}                                  & \textbf{65.00}                & 52.60                         & 31.80                                  & \multicolumn{1}{c|}{21.10}                                  & \multicolumn{1}{c|}{42.6±7.2}                                  & \cellcolor[HTML]{EFEFEF}44.4±8.1          & 55.30                         & \multicolumn{1}{c|}{\textbf{56.80}}                & \cellcolor[HTML]{EFEFEF}56.1±7.3          \\
            F+B+A+T+G & 47.60                         & 65.00                         & 36.60                                  & \multicolumn{1}{c|}{42.70}                                  & \multicolumn{1}{c|}{48.0±2.4}                                  & 46.30                         & \textbf{66.90}                & 28.40                                  & \multicolumn{1}{c|}{30.90}                                  & \multicolumn{1}{c|}{43.1±4.3}                                  & \cellcolor[HTML]{EFEFEF}45.5±6.3          & \textbf{66.70}                & \multicolumn{1}{c|}{48.20}                         & \cellcolor[HTML]{EFEFEF}57.4±9.5          \\
            Full+Dynamic   & \cellcolor[HTML]{EFEFEF}49.60 & \cellcolor[HTML]{EFEFEF}61.90 & \cellcolor[HTML]{EFEFEF}\textbf{38.40} & \multicolumn{1}{c|}{\cellcolor[HTML]{EFEFEF}\textbf{52.80}} & \multicolumn{1}{c|}{\cellcolor[HTML]{EFEFEF}\textbf{50.7±3.6}} & \cellcolor[HTML]{EFEFEF}47.20 & \cellcolor[HTML]{EFEFEF}59.70 & \cellcolor[HTML]{EFEFEF}\textbf{40.50} & \multicolumn{1}{c|}{\cellcolor[HTML]{EFEFEF}\textbf{35.40}} & \multicolumn{1}{c|}{\cellcolor[HTML]{EFEFEF}\textbf{45.7±4.3}} & \cellcolor[HTML]{EFEFEF}\textbf{48.2±6.5} & \cellcolor[HTML]{EFEFEF}65.50 & \multicolumn{1}{c|}{\cellcolor[HTML]{EFEFEF}53.30} & \cellcolor[HTML]{EFEFEF}\textbf{59.4±6.8} \\ \hline
            \end{tabular}
		\footnotesize \textit{Note: F: Face Feature, B: Body Feature, A: Audio Feature, T: Text Feature. G: Graph Network. We use recall (in \%) for each class and UAR (in \%) for overall performance.}
	\end{threeparttable}
	\label{Table: 3}
	\vspace{-1.0em}
\end{table*}

\begin{table}[h!]\footnotesize
    \vspace{-0.5em}
    \caption{Experimental results of dual attention ablation in GNN.}
    \vspace{-0.5em}
    \centering
    \def\temptablewidth{0.2\textwidth}
    \tabcolsep=0.35cm
    \renewcommand\arraystretch{1.5}
    \begin{threeparttable}
        \begin{tabular}{c|ccc}
            \hline
            \textit{\textbf{}} & NoXi-I                                    & NoXi-J                                    & UDIVA                                     \\ \hline
            No Node Att        & 48.0±3.4                                  & 44.2±5.3                                  & 57.5±7.0                                  \\
            No Edge Att        & 28.7±10.2                                 & 28.6±8.2                                  & 50.8±2.2                                  \\
            All                & \cellcolor[HTML]{EFEFEF}\textbf{50.7±3.6} & \cellcolor[HTML]{EFEFEF}\textbf{45.7±4.3} & \cellcolor[HTML]{EFEFEF}\textbf{59.4±6.8} \\ \hline
            \end{tabular}
        \footnotesize \textit{Note: We use UAR (in \%) for overall performance.}
    \end{threeparttable}
    \label{Table: 4}
    \vspace{-1.0em}
\end{table}

\subsubsection{Ablation of Node-Edge Attention} 
Table \ref{Table: 4} quantifies the need for dual attention mechanisms. Removing node attention (`No Node Att') results in moderate performance degradation (NoXi-I: 48.0\% vs. 50.7\%; NoXi-J: 44.2\% vs. 45.7\%; UDIVA: 57.5\% vs. 59.4\%), highlighting the importance of adaptive modality weighting. The strong decrease in the `No Edge Att' configurations (NoXi-I: 28.7\%; NoXi-J: 28.6\%; UDIVA: 50.8\%) emphasizes the critical role of edge attention in modeling asymmetric cross-modal dependencies. Optimal performance is achieved when node attention and edge attention are combined (`All').

\subsubsection{Ablation of Periodic Temporal Signals}
\begin{figure}[h!]
    \vspace{-1.0em}
	\setlength{\abovecaptionskip}{0.cm}
	\setlength{\belowcaptionskip}{-0.cm}
	\centering
	\includegraphics[width=2.2in]{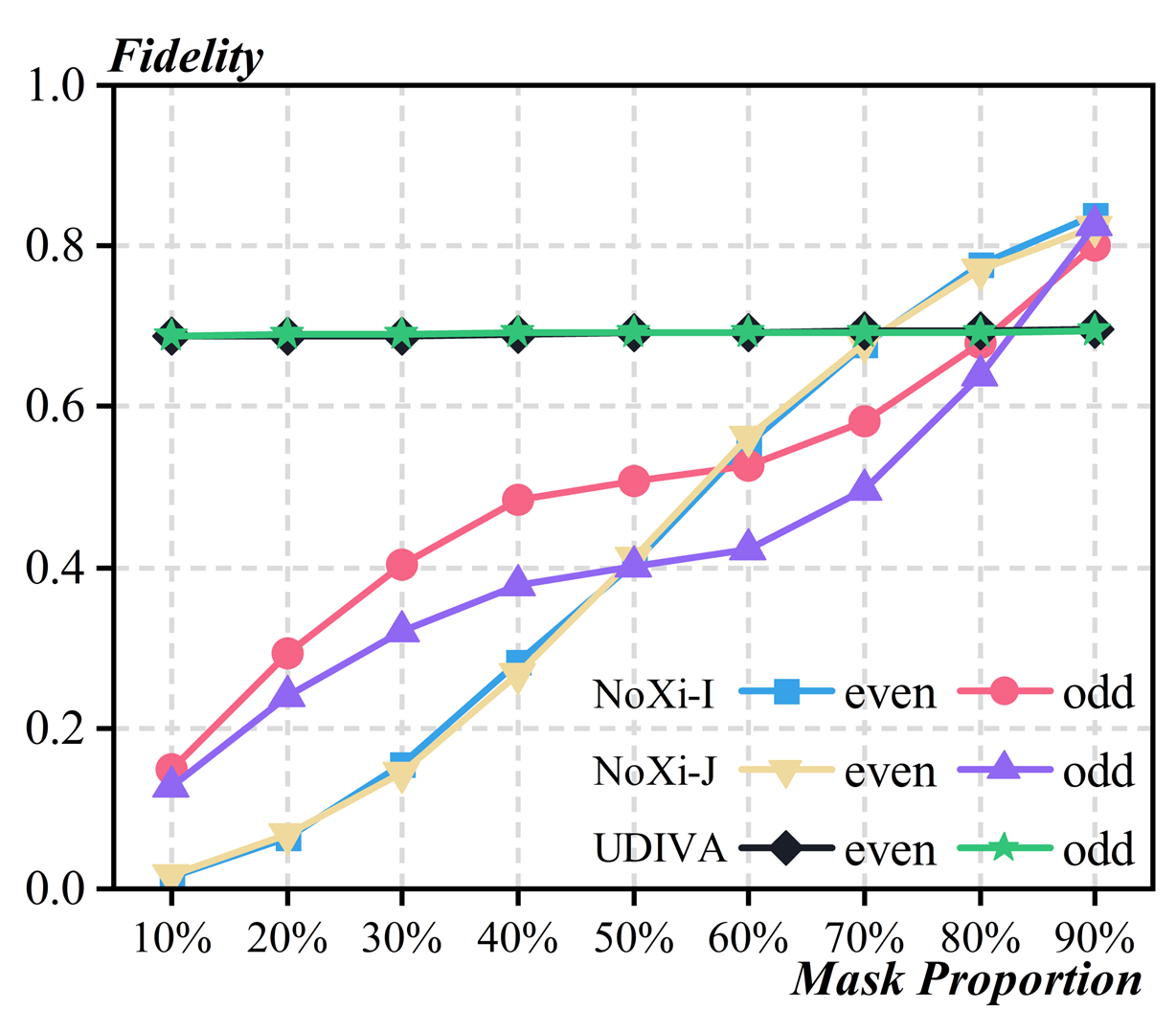}
	\vspace{-0.5em}
	\caption{Experiments on even and odd temporal signal masking.\label{Fig: 6}}
    \vspace{-1.5em}
\end{figure}

To evaluate the proposed periodic temporal modeling, we independently masked odd and even temporal signals at ratios ranging from 10\% to 90\%. The results (Fig. \ref{Fig: 6}) show that NoXi-I and NoXi-J exhibit consistent performance degradation with increasing masking ratios, as reflected by increasing fidelity distances. Odd and even signals are equally critical, with impairment of either leading to increased prediction bias. In contrast, UDIVA shows significant biases when either odd or even signals are masked, regardless of the masking ratio. This suggests that periodic temporal signals play an important role in modeling both asymmetric and symmetric relationships and highlights the importance of periodic temporal modeling in capturing dynamic relationship patterns.

\subsection{Hierarchical Dyadic Relationship Classification Analyses} \label{subsec: HDRCA}
\begin{figure}[h!]
    \vspace{-1.0em}
    \setlength{\abovecaptionskip}{0.cm}
    \setlength{\belowcaptionskip}{-0.cm}
    \centering
    \includegraphics[width=3.4in]{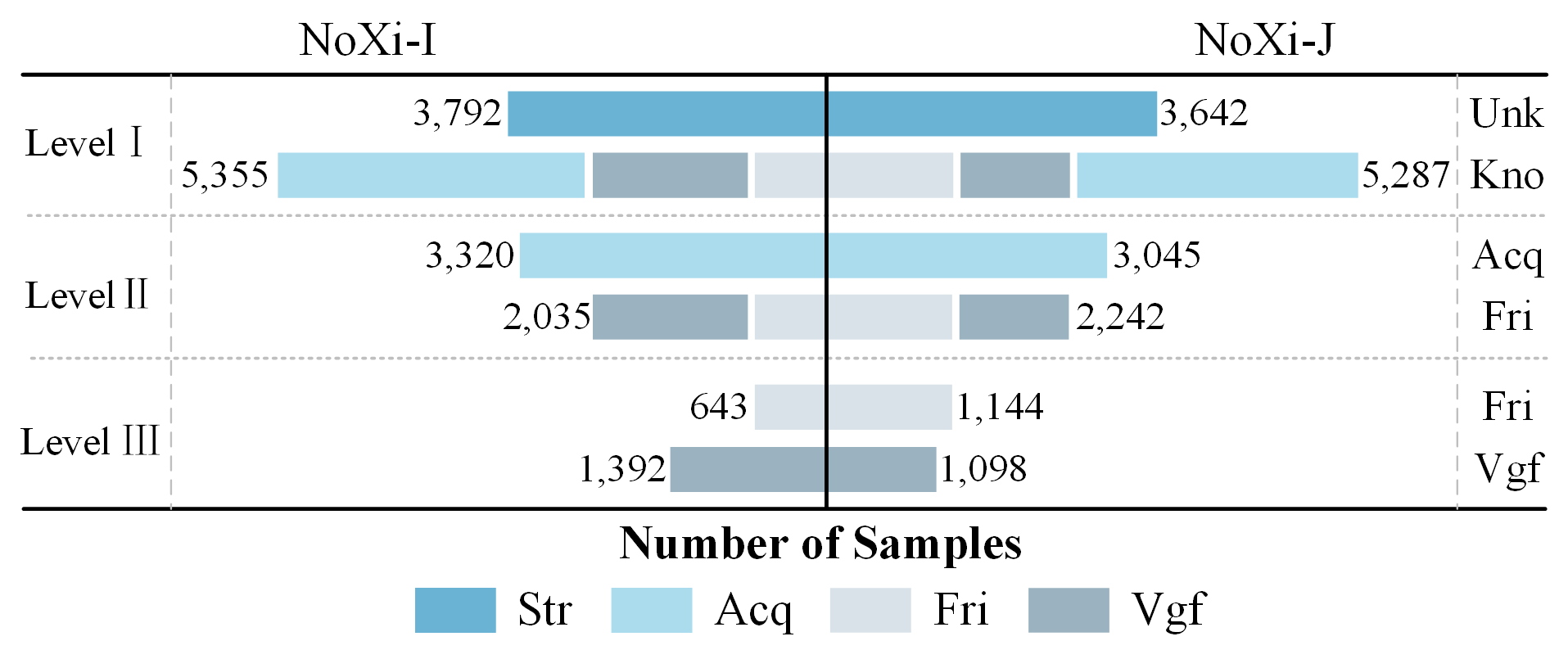}
    \vspace{-0.5em}
    \caption{Hierarchical data distribution on the NoXi dataset. \label{Fig: 7}}
\end{figure}

The pronounced class imbalance in the NoXi dataset (Table~\ref{Table: 1}) hinders the effective analysis of how multimodal cues, interaction dynamics, and temporal signals contribute to relationship recognition. To address this limitation, we perform hierarchical ablation experiments by restructuring the original 4-class task into three progressively refined levels (Fig. \ref{Fig: 7}): Level I Unknown (stranger) vs. Known (acquaintance/friend/very good friend). Level II: Acquaintance vs. Friend (including very good friend). Level III: Friend vs. Very Good Friend. Each level contains relatively balanced data distributions, and we retrain and retest each level independently.

\subsubsection{Hierarchical Relationship Classification Performance}
As shown in Table \ref{Table: 5}, hierarchical relationship classification exhibits different performance patterns at different levels of granularity. At Level I, the framework demonstrates robust discrimination between ``unknown'' and ``known'' relationships, achieving UAR values of 76.8\% (NoXi-I) and 74.6\% (NoXi-J), with elevated recall for ``known'' relationships (85.6\% NoXi-I, 87.0\% NoXi-J). This suggests that coarse-grained behavioural cues, such as formal greetings, provide sufficient discriminative power for binary classification. Performance declines at Level II (UAR: 70.3\% NoXi-I, 65.6\% NoXi-J), particularly in identifying ``friend'' (62.4\% NoXi-I, 67.4\% NoXi-J), likely due to semantic overlap in behavioural signatures (e.g., informal verbal exchange, sporadic eye contact) that often occurs between ``acquaintances'' and ``friends''. Conversely, Level III restores discriminability, achieving peak performance on NoXi-I (UAR: 89.8\%; very good friend recall: 94.1\%) and modest gains on NoXi-J (UAR: 68.7\%). This could be due to intimate behaviours such as synchronized emotional expression, infectious laughter, conversational depth, and prolonged mutual gazing, which tend to occur more frequently between close friends~\cite{93,94}.

\begin{table}[h!]\footnotesize
    \caption{Results of hierarchical classification on the NoXi.}
    \vspace{-0.5em}
    \centering
    \def\temptablewidth{0.2\textwidth}
    \tabcolsep=0.15cm
    \renewcommand\arraystretch{1.5}
    \begin{threeparttable}
        \begin{tabular}{c|ccc|ccc}
        \hline
        \textbf{}                                                   & \multicolumn{3}{c|}{NoXi-I}                               & \multicolumn{3}{c}{NoXi-J}\\ \hline
        {\color[HTML]{404040} }                                     & {\color[HTML]{404040} Unk.}                              & {\color[HTML]{404040} Kno.}                              & UAR  & Unk.  & Kno.    & UAR  \\ \cline{2-7} 
        \multirow{-2}{*}{{\color[HTML]{404040} \textit{Level I}}}   & \cellcolor[HTML]{EFEFEF}{\color[HTML]{404040} 67.9} & \cellcolor[HTML]{EFEFEF}{\color[HTML]{404040} 85.6} & \cellcolor[HTML]{EFEFEF}76.8±18.4 & \cellcolor[HTML]{EFEFEF}62.3 & \cellcolor[HTML]{EFEFEF}87.0 & \cellcolor[HTML]{EFEFEF}74.6±15.9 \\ \hline
        {\color[HTML]{404040} }                                     & {\color[HTML]{404040} Acq.}                              & {\color[HTML]{404040} Fri.}                              & UAR   & Acq.  & Fri. & UAR \\ \cline{2-7} 
        \multirow{-2}{*}{{\color[HTML]{404040} \textit{Level II}}}  & \cellcolor[HTML]{EFEFEF}{\color[HTML]{404040} 78.3}  & \cellcolor[HTML]{EFEFEF}{\color[HTML]{404040} 62.4} & \cellcolor[HTML]{EFEFEF}70.3±10.1 & \cellcolor[HTML]{EFEFEF}63.9  & \cellcolor[HTML]{EFEFEF}67.4  & \cellcolor[HTML]{EFEFEF}65.6±9.2  \\ \hline
        {\color[HTML]{404040} }                                     & {\color[HTML]{404040} Fri.}                              & {\color[HTML]{404040} Vgf.}                              & UAR  & Fri.  & Vgf.    & UAR  \\ \cline{2-7} 
        \multirow{-2}{*}{{\color[HTML]{404040} \textit{Level III}}} & \cellcolor[HTML]{EFEFEF}{\color[HTML]{404040} 85.5} & \cellcolor[HTML]{EFEFEF}{\color[HTML]{404040} 94.1}  & \cellcolor[HTML]{EFEFEF}89.8±8.3  & \cellcolor[HTML]{EFEFEF}71.8 & \cellcolor[HTML]{EFEFEF}65.7 & \cellcolor[HTML]{EFEFEF}68.7±5.2  \\ \hline
        \end{tabular}
        \footnotesize \textit{Note: We use recall (in \%) for each class and UAR (in \%) for overall performance.}
    \end{threeparttable}
    \label{Table: 5}
    \vspace{-0.5em}
\end{table}

\begin{figure*}[t!]
	\setlength{\abovecaptionskip}{0.cm}
	\setlength{\belowcaptionskip}{-0.cm}
	\centering
	\includegraphics[width=5.5in]{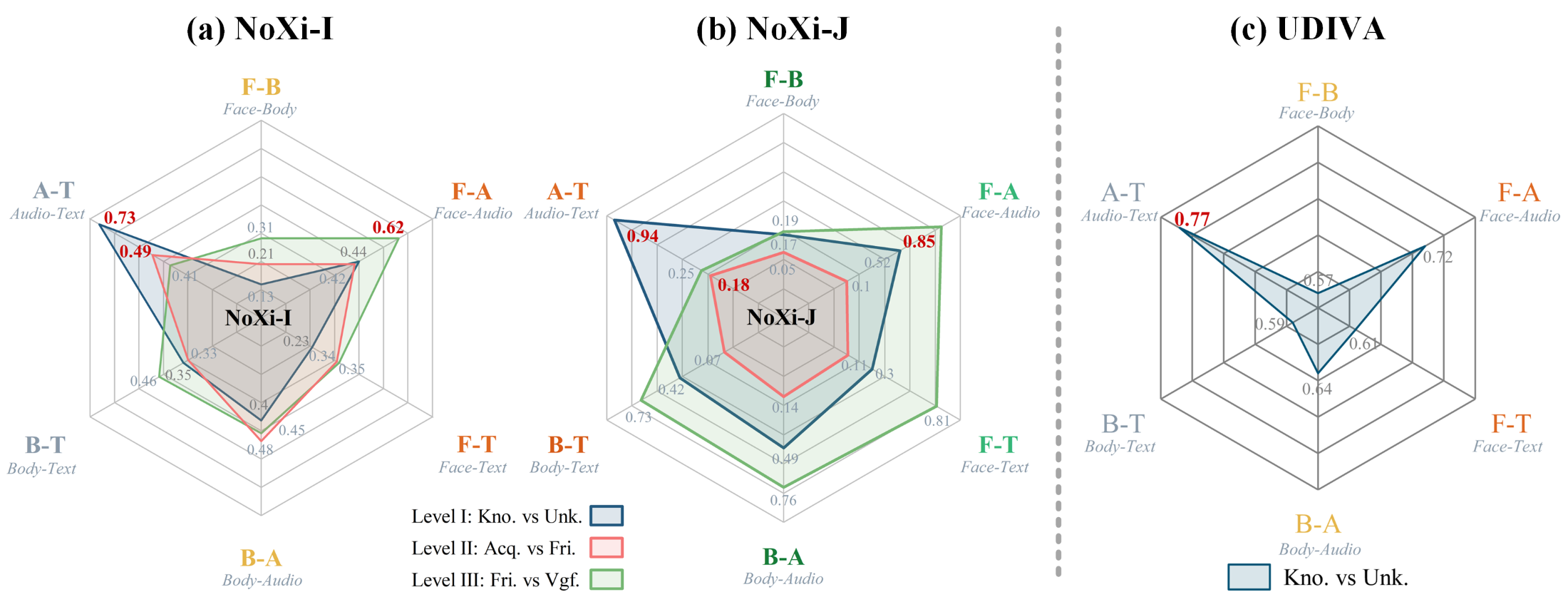}
	\caption{Experiments on masking multimodal interactions in graph networks. Specifically, we mask edges that represent interactions of different modalities. These include edges between audio and text (A-T), body and text (B-T), body and audio (B-A), face and text (F-T), face and audio (F-A), and face and body (F-B). The results are evaluated using fidelity metrics, with higher values indicating more important ones. \label{Fig: 8}}
    \vspace{-1.0em}
\end{figure*}

\subsubsection{Dyadic Interaction Ablation Study}
To investigate the factors that contribute to relationship recognition, we conducted an interaction ablation study by masking multimodal information during dyadic interactions. As shown in Fig. \ref{Fig: 8}, Level I demonstrates that audio-text (A-T) interactions achieve the highest fidelity (NoXi-I: 0.73, NoXi-J: 0.94, UDIVA: 0.77), indicating that the model relies on voice and speech content (text) to distinguish ``\textit{strangers}'' from ``\textit{acquaintances}''. In Level II, the model expands its focus to include A-T, body-audio (B-A), body-text (B-T), and face-audio (F-A) interactions with comparable fidelity, reflecting the overlapping behavioural characteristics between ``\textit{acquaintances}'' and ``\textit{friends}'' that require richer multimodal and specifically nonverbal behavioural data for differentiation. At Level III, face-audio (F-A) interactions become critical for distinguishing friends from very good friends (NoXi-I: 0.62, NoXi-J: 0.85), as both primarily encode nonverbal behaviour and expressive cues. These cues help distinguish the subtle differences between ``\textit{friends}'' and ``\textit{very good friends}''.

\subsubsection{Temporal Signal Ablation Study}  
\begin{figure*}[t!]
	\setlength{\abovecaptionskip}{0.cm}
	\setlength{\belowcaptionskip}{-0.cm}
	\centering
	\includegraphics[width=5.4in]{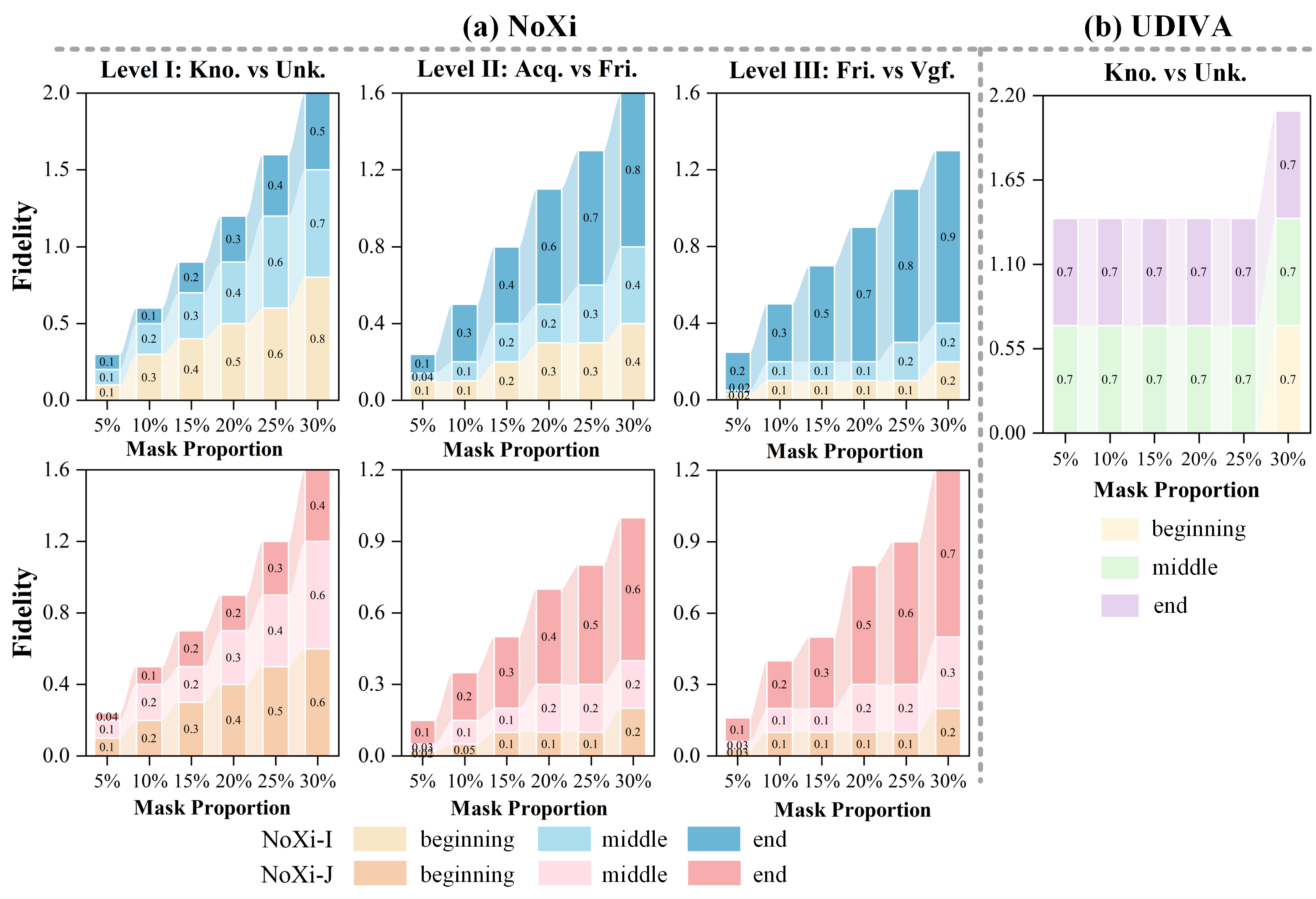}
	\caption{Temporal signal masking experiments in which we selectively mask clips at the beginning, middle, and end of the clip sequence. The masking percentage is from 10\% to 30\% of the total video duration. For example, when the beginning segment is masked, the middle and end segments remain visible. \label{Fig: 9}}
    \vspace{-1.5em}
\end{figure*}

To evaluate the importance of temporal signals, we selectively masked the clips from the beginning, middle, or end of videos. The results in Fig. \ref{Fig: 9} show that as more temporal information is obscured, the model's predictions are more negatively affected. For NoXi, beginning and middle segments are critical for Level I ``\textit{unknown}'' and ``\textit{known}'' recognition, as early interactions (e.g., opening remarks) and mid-segment exchanges (e.g., introductory conversations) provide sufficient cues, while UDIVA relies more on middle and end segments. At finer levels (Level II and III) in NoXi, the model shifts focus to middle and end signals, which contain nuanced behavioural cues and deeper nonverbal and expressive exchanges necessary to distinguish ``\textit{acquaintances}'', ``\textit{friends}'', and ``\textit{very good friends}''. In contrast, earlier segments dominated by superficial interactions (e.g., greetings) are less informative, underscoring the importance of mid-to-late temporal signals in capturing fine-grained social dynamics.

\begin{figure*}[h!]
	\setlength{\abovecaptionskip}{0.cm}
	\setlength{\belowcaptionskip}{-0.cm}
	\centering
	\includegraphics[width=6.2in]{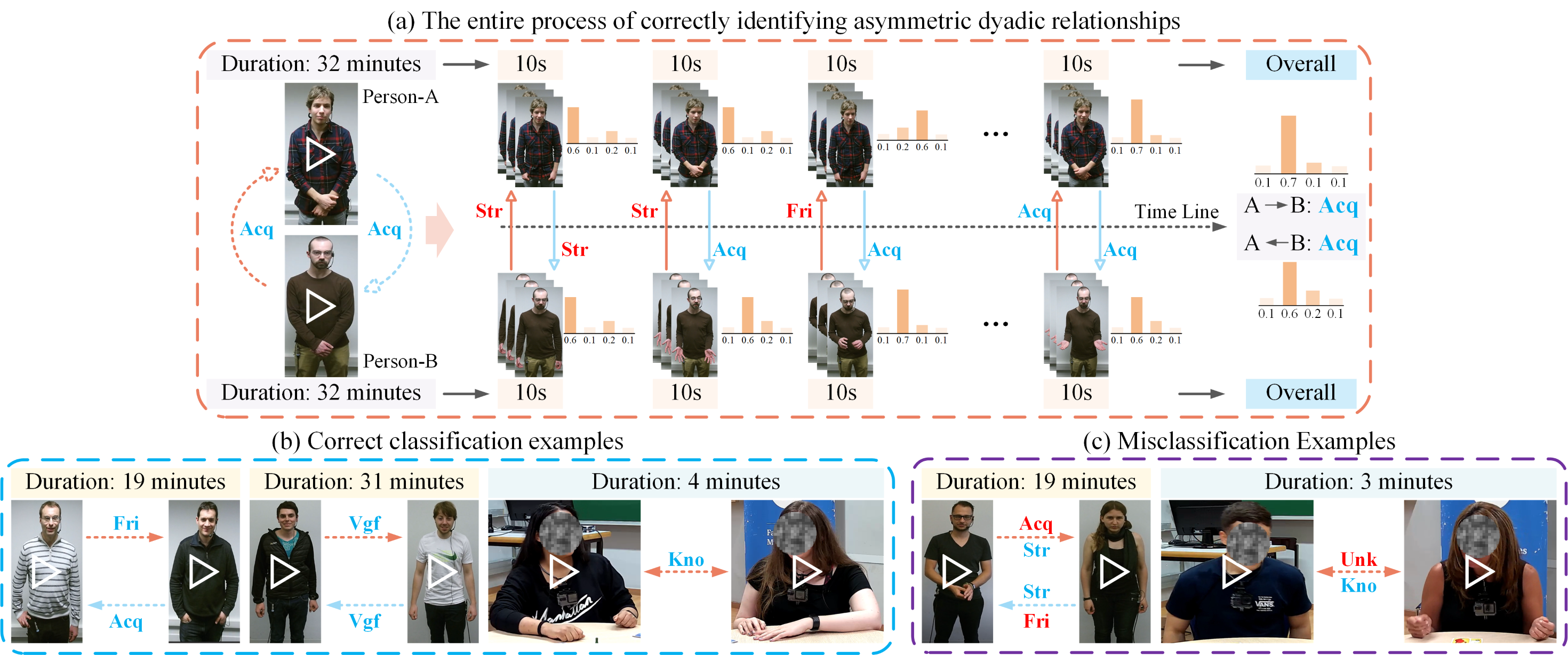}
	\caption{Qualitative Analysis.  (a) Illustration of the entire process of asymmetric relationship recognition.  (b) Examples of correct relationship recognition.  (c) Examples of incorrect relationship recognition. Screenshots of standing people are taken from the NoXi database, collected for the ARIA-VALUSPA project~\cite{49}. Screenshots of sitting people (their faces blurred for anonymity) are taken from the UDIVA V0.5 dataset~\cite{48}, collected in the scope of the research project entitled ``Understanding Face-to-Face Dyadic Interactions through Social Signal Processing. \textit{Note that the text in red represents false predictions, and the blue represents the ground truth relationships.} \label{Fig: 10}}
    \vspace{-1.0em}
\end{figure*}

\subsection{Qualitative Analyses}
Fig. \ref{Fig: 10} provides examples from the NoXi (standing interactions) and UDIVA (sitting interactions) datasets. Fig. \ref{Fig: 10}(a) depicts the clip-level relationship learning process. The framework segments a video of Person A and Person B, who perceive each other as ``\textit{acquaintances}'', into clips. For each clip, the model captures asymmetric bidirectional interaction patterns to infer relationships, dynamically updating predictions (e.g., from unacquainted to acquainted) as the video progresses. Local clip-level predictions are integrated with global temporal signals to refine the final relationship recognition by correcting intermediate errors and enhancing robustness.

Fig. \ref{Fig: 10}(b) and (c) show correct and incorrect classifications. The model accurately identifies bidirectional or unidirectional relationships across both datasets, demonstrating its robustness. However, we observe that misclassifications predominantly occur in mixed-gender interactions (male-female), which are underrepresented in the datasets compared to same-gender pairs (male-male, female-female). Same-gender interactions exhibit distinct patterns (e.g., lower-frequency audio in male-male, higher-pitched audio, and specific facial features in female-female), which the model could learn effectively due to their prevalence. In contrast, mixed-gender interactions introduce variability, which the model may interpret as noise, reducing robustness. These results highlight the need for greater dataset diversity to improve generalization across gender-diverse interactions.

\section{Discussion, Conclusion and Future Work}

In this paper, we propose AsyReC, a multimodal graph-based framework for spatio-temporal asymmetric dyadic relationship classification. Our framework introduces three main contributions: (1) a triplet graph neural network, enhanced by node-edge dual attention mechanisms, that dynamically models asymmetric perceptual cues through adaptive weighting of multimodal features; (2) a clip-level dyadic relationship learning paradigm that captures the temporal evolution of social interactions, enabling fine-grained analysis of real-world behavioral dynamics; and (3) a periodic temporal encoding method that maps clip indices onto sine/cosine waveforms to explicitly capture recurrent behavioural patterns, addressing the fragmented temporal modeling of conventional approaches. Extensive experiments on the NoXi and UDIVA datasets demonstrate the state-of-the-art performance of the proposed framework, while ablation studies confirm the need for asymmetric interaction modeling and temporal periodicity encoding. 

The hierarchical relationship recognition exploration introduced in this study holds significant promise for real-time interactive systems requiring efficient social perception. Experimental results in Sec.~\ref{subsec: HDRCA} show coarse-grained relationship classification distinguishing between ``known'' and ``unknown'' interactions, and this can potentially be detected with lightweight models for the initial processing stages of time-sensitive applications such as public service platforms or live customer support interfaces~\cite{98}. These systems can prioritize urgent cases or unknown users through efficient binary classification, reserving more data-intensive analysis for later, more nuanced relationship evaluations.

As shown in Fig. \ref{Fig: 8}, systems with hardware constraints, such as those lacking high-resolution cameras, could prioritize audio-text fusion for basic relationship screening. Conversely, applications emphasizing emotional reciprocity (e.g., companionship-oriented interfaces or mental health support tools) might focus on face-audio modalities to detect trust-building signals, even with sparse training data. These findings empower developers to tailor multimodal configurations to specific use-case requirements—such as privacy, latency, or deployment scale—without compromising robustness.

The temporal ablation studies (Fig. \ref{Fig: 9}) reveal opportunities for optimizing computational efficiency in resource-constrained embedded systems. For instance, recognizing acquaintances or strangers can leverage early-to-mid temporal segments, while detecting close friendships benefits from later segments. By selectively processing relevant temporal windows, real-time systems could reduce inference latency while maintaining accuracy. Such strategies are particularly valuable for edge-deployed platforms, where hardware limitations conflict with real-time demands.

Future work on the AsyReC framework can be focused on several key directions to advance the field of social relationship recognition. For example, dataset diversity can be enhanced by incorporating mixed-gender and cross-cultural interaction scenarios, as well as by extending to complex multi-person interactions. This extension aims to address under-representation issues and improve the generalizability of the model across different social contexts. Additionally, the exploration of explainable methods can address the inherent ``black box'' nature of deep learning models in social relationship recognition. By elucidating how machines learn and identify human social interaction patterns, these methods can significantly improve the interpretability and reliability of socially intelligent systems, fostering greater trust and applicability in real-world scenarios.

\bibliographystyle{IEEEtran}
\bibliography{IEEEabrv,Bibliography}


\vspace{-3.5em}
\begin{IEEEbiographynophoto}{Wang Tang} is currently working towards the Ph.D. at the College of Electronics and Information Engineering, Sichuan University, Chengdu, China. He is a visiting scholar at the AFAR in the Department of Computer Science and Technology, University of Cambridge.
\end{IEEEbiographynophoto}
\vspace{-3.5em}

\begin{IEEEbiographynophoto} {Fethiye Irmak Dogan} is a Postdoctoral Researcher in the AFAR Lab at the University of Cambridge (UK). She received her Ph.D. degree in Computer Science from KTH Royal Institute of Technology (Sweden). Her research focuses on human-robot interaction, robot learning, and explainability.
\end{IEEEbiographynophoto}
\vspace{-3.5em}

\begin{IEEEbiographynophoto}{Linbo Qing} is currently a Professor with the College of Electronics and Information Engineering, Sichuan University. His main research interests include artificial intelligence and computer vision, image processing, visual computing, data mining, and digital health.
\end{IEEEbiographynophoto}
\vspace{-3.5em}

\begin{IEEEbiographynophoto}{Hatice Gunes} is a Full Professor of Affective Intelligence and Robotics (AFAR) in the Department of Computer Science and Technology, University of Cambridge, leading the \href{https://cambridge-afar.github.io/} {Cambridge AFAR Lab}. She is a former President of the Association for the Advancement of Affective Computing, a former Faculty Fellow of the Alan Turing Institute, and is currently a Fellow of the EPSRC and a Staff Fellow of Trinity Hall.
\end{IEEEbiographynophoto}

\end{document}